\pdfoutput=1
\documentclass[11pt]{article}

\usepackage[]{acl}

\usepackage{times}
\usepackage{latexsym}

\usepackage[T1]{fontenc}
\usepackage[utf8]{inputenc}
\usepackage{helvet}

\usepackage{microtype}

\usepackage{inconsolata}

\usepackage{booktabs}
\usepackage{multirow}
\usepackage{adjustbox}
\usepackage{amsmath}
\usepackage{color}

\usepackage{lipsum}

\usepackage[linesnumbered,ruled,vlined]{algorithm2e}
\usepackage{algpseudocode}

\definecolor{fontgray}{RGB}{44, 62, 80}
\definecolor{myred}{RGB}{235, 47, 6} %rgb()
\definecolor{summertime}{RGB}{245, 205, 121}
\definecolor{darkgrass}{RGB}{0, 148, 50}
\definecolor{myblue}{RGB}{0, 168, 255}
\definecolor{mygray}{RGB}{158, 158, 158}
\definecolor{puffin}{RGB}{250, 152, 58}
\definecolor{lowpurple}{RGB}{210, 180, 222}
\definecolor{lowblue}{RGB}{102,178,255}
\definecolor{lowred}{RGB}{245, 183, 177}
\definecolor{deeppurple}{RGB}{142, 68, 173}
\definecolor{nephritis}{RGB}{39, 174, 96}

\definecolor{deepblue}{RGB}{41, 128, 185}
\definecolor{shymoment}{RGB}{162, 155, 254}
\definecolor{firstdate}{RGB}{250, 177, 160}
\definecolor{mintleaf}{RGB}{0, 184, 148}
\definecolor{alizarin}{RGB}{231, 76, 60}
\definecolor{soaring}{RGB}{149, 175, 192}
\definecolor{electronblue}{RGB}{9, 132, 227}
\definecolor{pinkgla}{RGB}{0, 184, 148}
\definecolor{coral}{RGB}{255, 127, 80}

\newcommand\rurl[1]{\href{https://#1}{\nolinkurl{#1}}}

\newlength\mylen

\usepackage{amsmath,amssymb}
\usepackage{subcaption}
\usepackage{threeparttable}
\usepackage{tikz}
\usepackage{pgf}
\usepackage{pgfplots}
\pgfplotsset{compat=1.18}
\usetikzlibrary{patterns}
\usepackage{tikz-qtree}
\usetikzlibrary{arrows,decorations.pathmorphing,backgrounds,positioning,fit,petri,shapes.misc, arrows.meta,shapes.geometric,decorations.markings,calc,shadows.blur,decorations.pathreplacing,quotes,matrix,shapes.symbols}
\usetikzlibrary{external}
\tikzset{
table/.style={
first column text width/.code={%
	\tikzset{
		column 1/.style={
			nodes={text width=##1, font=\bfseries}
		},
	}
},
first column text width=5em
}
}

\usepackage{arydshln}
\newcommand{\norm}[1]{\left\lVert#1\right\rVert}

\newcommand{\squishlist}{
	\begin{list}{$\bullet$}
		{ \setlength{\itemsep}{0pt}
			\setlength{\parsep}{3pt}
			\setlength{\topsep}{3pt}
			\setlength{\partopsep}{0pt}
			\setlength{\leftmargin}{1.5em}
			\setlength{\labelwidth}{1em}
			\setlength{\labelsep}{0.5em} } }

	\newcounter{Lcount}
	\newcommand{\squishlisttwo}{
		\begin{list}{\arabic{Lcount}. }
			{ \usecounter{Lcount}
				\setlength{\itemsep}{0pt}
				\setlength{\parsep}{0pt}
				\setlength{\topsep}{0pt}
				\setlength{\partopsep}{0pt}
				\setlength{\leftmargin}{2em}
				\setlength{\labelwidth}{1.5em}
				\setlength{\labelsep}{0.5em} } }
		
		\newcommand{\squishend}{
	\end{list} }

\usepackage[most]{tcolorbox}
\usepackage{tikz}
\lstdefinestyle{python}{
    language=Python,
    basicstyle=\fontsize{7}{8}\ttfamily,
    keywordstyle=\color{blue},
    commentstyle=\color{gray},
    stringstyle=\color{black},
    showstringspaces=false,
    breaklines=true,
    breakindent=0pt,
    breakatwhitespace=false,
    escapeinside={(*@}{@*)}
}
\lstdefinestyle{plain}{
    basicstyle=\fontsize{8}{10}\ttfamily,
    keywordstyle=\color{blue},
    commentstyle=\color{gray},
    stringstyle=\color{green},
    showstringspaces=false,
    breaklines=true,
    breakatwhitespace=false,
    breakindent=0pt,
    escapeinside={(*@}{@*)}
}
\title{\textsc{ReFT}: Reasoning with \textsc{Re}inforced \textsc{F}ine-\textsc{T}uning}
\author{Trung Quoc Luong$^{*}$, Xinbo Zhang$^{*}$, Zhanming Jie\thanks{~indicates equal contribution, $\dagger$ indicates corresponding author}, Peng Sun$^\dagger$, Xiaoran Jin, Hang Li \\
  ByteDance Research \\
  \texttt{\{trung.luong, zhangxinbo.freya, allan.jie\}@bytedance.com} \\
  \texttt{\{wanhesong, xiaoran.jin, lihang.lh\}@bytedance.com} \\
  }

\begin{document}
\maketitle

\begin{abstract}
One way to enhance the reasoning capability of Large Language Models (LLMs) is to conduct Supervised Fine-Tuning (SFT) using Chain-of-Thought (CoT) annotations. 
This approach does not show sufficiently strong generalization ability, however, because the training only relies on the given CoT data. 
In math problem-solving, for example, 
%there is usually one annotated valid path of reasoning for each question in the training data. 
there is usually only one annotated reasoning path for each question in the training data. 
%It would be better that the algorithm could learn from multiple valid paths of reasoning of each question. 
%It would be better that the algorithm could learn from multiple annotated paths of reasoning for each question. 
Intuitively, it would be better for the algorithm to learn from multiple annotated reasoning paths given a question.
To address this issue,  we propose a simple yet effective approach called \textit{Reinforced Fine-Tuning} (ReFT) to enhance the generalizability of learning LLMs for reasoning, with math problem-solving as an example.  
%ReFT uses the same set of CoT annotations for training as conventional SFT.
%ReFT uses the same set of questions and CoT annotations for training as conventional SFT.
%It first initializes the model in a similar way to SFT. 
ReFT first warmups the model with SFT, 
%It then employs an on-line reinforcement learning algorithm, specifically the PPO algorithm, to further fine-tune the model, where more paths of reasoning are automatically sampled given the question and the rewards are automatically derived from the answer. 
and then employs on-line reinforcement learning, specifically the PPO algorithm in this paper, to further fine-tune the model, 
where an abundance of reasoning paths are automatically sampled given the question and the rewards are naturally derived from the ground-truth answers. 
% Extensive experiments on GSM8K, MathQA, and SVAMP datasets show that ReFT significantly outperforms SFT with the same training data (CoT annotations), 
%Extensive experiments on GSM8K, MathQA, and SVAMP datasets show that ReFT significantly outperforms SFT with the same training data (questions and CoT annotations),
Extensive experiments on GSM8K, MathQA, and SVAMP datasets show that ReFT significantly outperforms SFT,
%indicating that ReFT has superior generalization ability. 
and the performance can be potentially further boosted by combining inference-time strategies such as majority voting and re-ranking. 
Note that ReFT obtains the improvement by learning from the same training questions as SFT, 
without relying on extra or augmented training questions.
This indicates a superior generalization ability for ReFT
\footnote{Code: \url{https://github.com/lqtrung1998/mwp_ReFT}}.
% \footnote{Code at \href{https://anonymous.4open.science/r/mwp_ReFT-6ECF}{\nolinkurl{r/mwp_ReFT-6ECF}}}.
%The performance of ReFT can be further improved when combined with different inference strategies such as majority voting and re-ranking.  
%The performance of ReFT can be further boosted by combining inference-time strategies such as majority voting and re-ranking. 
% Analysis also shows that ReFT indeed exhibits a strong tendency against over-fitting, compared to SFT.
% The code of this work is publicly available
% \footnote{Code at \url{https://anonymous.4open.science/r/mwp_ReFT-6ECF}}.
\end{abstract}

\section{Introduction}
\begin{figure}
    \centering
    \adjustbox{max width=1.0\linewidth}{
        \includegraphics{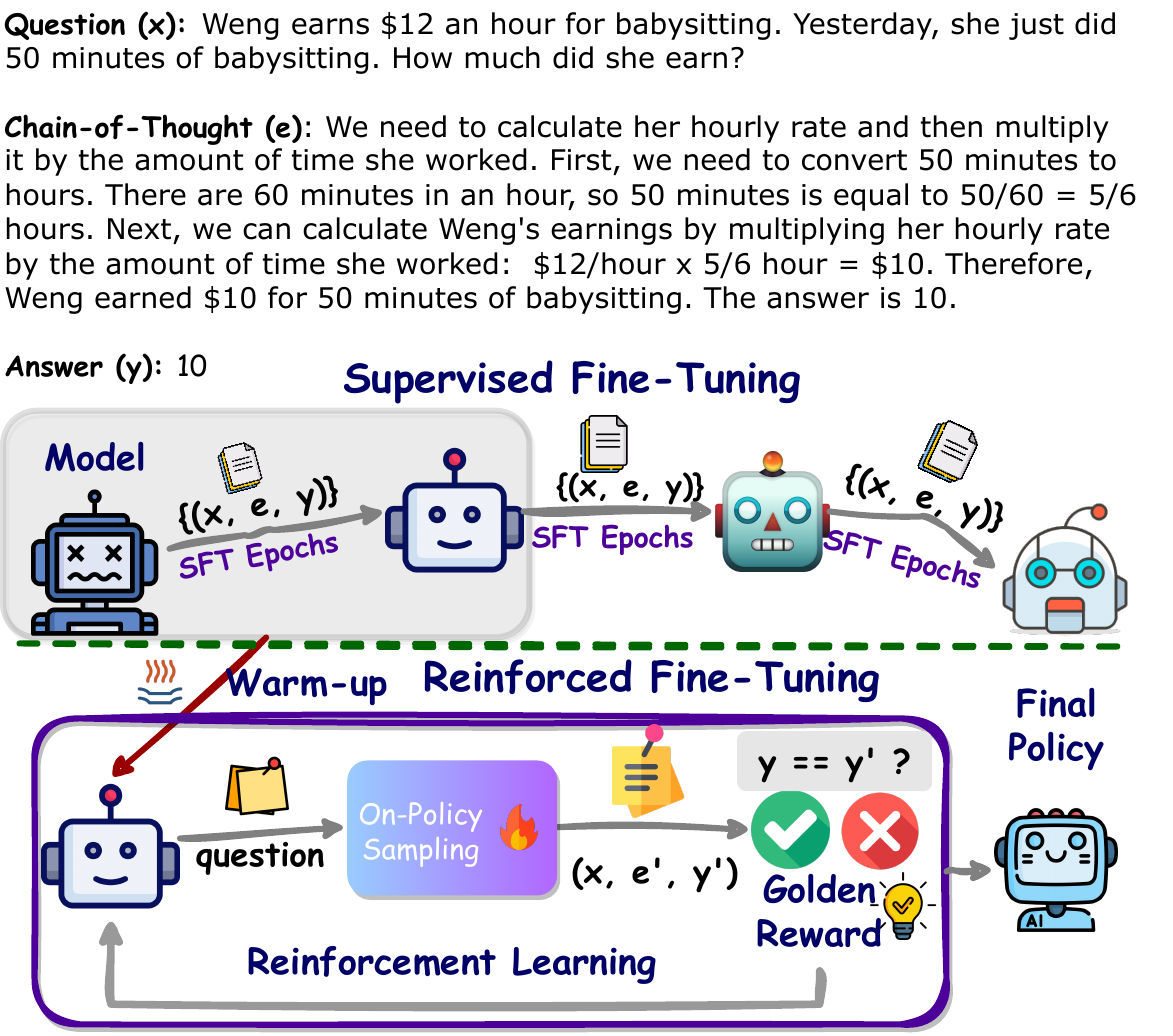}
    }
    \vspace*{-5mm}
    \caption{An example of question ($x$), CoT ($e$), and answer ($y$) in GSM8K~\cite{cobbe2021training}. 
    The SFT process iterates several epochs on the training data.
    The proposed ReFT warm-up from SFT and performs RL training on the same data.
    }
    \label{fig:sft_vs_rft}
\end{figure}

The state-of-the-art approaches to solving math problems~\cite{luo2023wizardmath,wang2023mathcoder} employ Supervised Fine-Tuning (SFT) to train the models using Chain-of-Thought (CoT) annotations~\cite{wei2022chain}. As shown in Figure \ref{fig:sft_vs_rft}, a CoT annotation outlines the intermediate reasoning steps toward solving a math problem. 

Usually there is one CoT annotation for each question in the training data, i.e., one correct reasoning path, which is utilized in SFT. We observe that this may result in relatively weak generalization abilities of the SFT models. It is often the case that multiple valid CoT annotations exist for the same question~\cite{cobbe2021training,zhang2023interpretable}, underscoring the need for a more powerful fine-tuning approach. To address this problem, we propose a simple yet effective approach called \textit{Reinforced Fine-Tuning} (ReFT) (Figure \ref{fig:sft_vs_rft} bottom).
% , depicted in the lower part of Figure \ref{fig:sft_vs_rft}.

\begin{figure}
    \centering
    \adjustbox{max width=1.0\linewidth}{
        \includegraphics{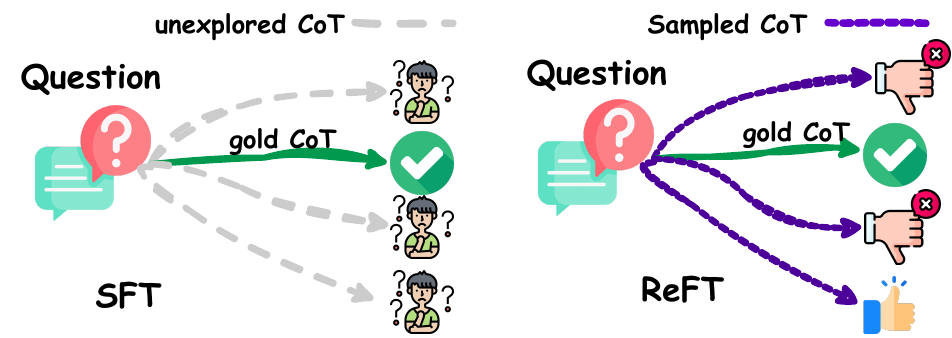}
    }
    \vspace*{-5mm}
    \caption{Comparison between SFT and ReFT on the presence of CoT alternatives.}
    \label{fig:sft_vs_ref_path}
\end{figure}

ReFT commences with a warm-up stage involving Supervised Fine-Tuning (SFT) in one or two epochs (Figure \ref{fig:sft_vs_rft}, shaded box). This initial stage equips the model with the ability to generate correct responses to mathematical problems to some extent, as demonstrated in prior work~\cite{cobbe2021training}. 
Next, ReFT proceeds to further refine the model through the utilization of an online Reinforcement Learning (RL) algorithm~\cite{sutton2018reinforcement}, specifically Proximal Policy Optimization (PPO)~\cite{schulman2017proximal} in this paper. 
%By employing PPO, 
In this way,
ReFT is able to sample multiple correct reasoning paths or CoT annotations and learn from them (Figure \ref{fig:sft_vs_ref_path}, right). 

%Since the training data includes the ground-truth answers, the rewards for executing the PPO algorithm can be naturally derived from them. 
Since the training data include ground-truth answers, the golden rewards can be naturally derived from them when training PPO.
Consequently, there is no requirement for a separately trained reward model.
In contrast, RLHF~\cite{ouyang2022training} has to utilize a reward model that is learned from human-labeled data.

During the warm-up stage, ReFT acquires a certain level of accuracy by supervised learning. In the RL stage, ReFT further enhances its ability by reinforcement learning through sampling various CoT reasoning paths.
In this way, ReFT gets much richer supervision signals than SFT.
This approach enables ReFT to greatly improve generalization in math problem-solving~\cite{gao2018reinforcement,brown2020better}. %as illustrated in Figure \ref{fig:sft_vs_ref_path}. 
%Note that ReFT outperforms SFT by using the same training questions as SFT.
% In contrast to RLHF~\cite{ouyang2022training} in the training of LLMs, which utilizes PPO to align with human-labeled data, ReFT leverages PPO to enhance the accuracy of math problem-solving using the same training data as SFT.
%In contrast to RLHF~\cite{ouyang2022training} which utilizes a reward model learned from human-labeled data, 
%ReFT adopts a golden reward derived from the ground-truth answer in the training data.
Note that ReFT outperforms SFT by using the same training questions,
without relying on extra or augmented training questions.
In fact, ReFT does not conflict with such data engineering and can be seamlessly combined with it.

% Our contributions can be summarized as follows:
Our contributions are as follows:
\squishlist

\item We introduce a novel fine-tuning approach, reinforced fine-tuning (ReFT), which utilizes reinforcement learning to solve math problems. ReFT exhibits enhanced generalization capabilities compared to conventional supervised fine-tuning when trained on the same dataset.

\item We conduct extensive experiments using two foundational models, CodeLLAMA~\cite{roziere2023code} and Galactica~\cite{taylor2022galactica}, on three standard datasets: GSM8K~\cite{cobbe2021training}, MathQA~\cite{amini2019mathqa}, and SVAMP~\cite{patel2021nlp}. Our experiments cover both natural language and program-based CoTs, demonstrating the significantly improved performance and generalization ability of ReFT.

\item Additionally, we demonstrate that ReFT benefits from both majority voting~\cite{wang2022self} and reward model reranking~\cite{uesato2022solving} at inference-time, further improving its performance.
\squishend

\section{Related Work}

\paragraph{Math Problem Solving}
Recent research efforts focus on CoT prompt design and data engineering. 
Most of them attempted to make CoT comprehensive and fine-grained to present the step-by-step reasoning solutions~\cite{nye2021show,fu2022complexity,zhou2022least,khot2022decomposed,zelikman2022star,imani2023mathprompter,miao2023selfcheck}.
\citet{gao2023pal} further proposed to use the Python program as CoT prompt, demonstrating more accurate reasoning steps and significant improvements over the natural language CoT~\cite{wei2022chain}.
\citet{zhou2023solving} introduced a prompting method that generates code to verify the intermediate reasoning step with GPT-4~\cite{openai2023gpt4}, thus achieving state-of-the-art performance on GSM8K~\cite{cobbe2021training} and MATH~\cite{hendrycks2021measuring}.  
Another line of work focuses on improving the quality of CoT~\cite{wang2023mathcoder,liu2023tinygsm,yu2023metamath} and increasing the amount of CoT data~\cite{luo2023wizardmath,yue2023mammoth} from OpenAI's ChatGPT (\texttt{gpt-3.5-turbo}) or GPT-4\footnote{\url{https://chat.openai.com/}}. 

\paragraph{Reinforcement Learning}
Our work is mostly related to the recent work that applies PPO~\cite{schulman2017proximal} to natural language process for aligning human preferences~~\cite{ouyang2022training}. 
Since then, several training algorithms have been proposed to efficiently improve the alignment,
including direct preference optimization (DPO)~\cite{rafailov2023direct}, identity preference optimization (IPO)~\cite{azar2023general}, and Kahneman-Tversky optimization (KTO)~\cite{ethayarajh2023halos}. 
Other than the purpose of alignment, we aim to adopt reinforcement learning as a fine-tuning paradigm to improve performance over conventional supervised fine-tuning.
% Our work is mostly related to reinforcement learning for (RLHF)~\cite{ouyang2022training} in natural language processing (NLP). 
% Several training algorithms including proximal policy optimization (PPO)~\cite{schulman2017proximal}, direct preference optimization (DPO)~\cite{rafailov2023direct}, identity preference optimization (IPO)~\cite{azar2023general}, and Kahneman-Tversky optimization (KTO)~\cite{ethayarajh2023halos}, have been proposed to improve the alignment with human preferences. 
% Unlike their purpose of alignment, we aim to adopt reinforcement learning as a fine-tuning paradigm to improve performance over conventional supervised fine-tuning. 
% \xinbo{Is the logic of this sentence right? Are these algorithms proposed for alignment?} -- modifications done.

Specifically for solving math problems, \citet{uesato2022solving} and \citet{lightman2023lets}  trained an outcome-based or process-based reward model to perform reranking~\cite{cobbe2021training} to achieve much better performance over SFT and majority voting~\cite{wang2022self}. 
While our approach aims to improve the performance of the policy itself, these reward model reranking approaches can be easily integrated into the resulting policy model.

\begin{algorithm*}[t!]
% \scriptsize
% \algsetup{linenosize=\}
\newcommand{\mycommfont}[1]{\normalfont\textcolor{black}{#1}}
\SetCommentSty{mycommfont}
\DontPrintSemicolon
\KwIn{$\mathcal{D}_{train} = \{(\boldsymbol{x}, \boldsymbol{e}, \boldsymbol{y})\}$: Tuples of (\textit{question}, \textit{CoT}, \textit{answer}), $W$: number of warm-up steps, $T$: number of RL steps, $U$: number of updates per RL step, $\boldsymbol{\pi}_\theta^{(0)}$: Initial policy.}
% \xinbo{no notation before for $\boldsymbol{x}, \boldsymbol{r}, \boldsymbol{y}$. I think there is a bit confusing in this whole section3 for $\boldsymbol{x}, \boldsymbol{r}, \boldsymbol{y}$} -- modifications done

\KwOut{$\boldsymbol{\pi}_{\boldsymbol{\theta}}$: Final policy}

$\boldsymbol{\pi}_{\boldsymbol{\theta}}=\boldsymbol{\pi}_{\boldsymbol{\theta}}^{(0)}$ \;
//~\textcolor{mintleaf}{\em Warm-up stage}\;
%\tcp{Update model gradients}
\For{$i\gets 1$ \KwTo $W$}{
%	Optimize $\boldsymbol{\pi}_\theta$ on $\mathcal
    $\boldsymbol{x}, \boldsymbol{e}, \boldsymbol{y} \sim  \mathcal{D}_{train}$ \tcp*[r]{\textcolor{mintleaf}{Sample mini-batch from $\mathcal{D}_{train}$}}
    % $\mathcal{M} \gets \mathcal{M} - lr \nabla L_{sft}(x, CoT)$\;
    $\boldsymbol{\theta} = \textsc{Optimization\_Step}(\mathcal{L}_{SFT}(\boldsymbol{\theta}))$ \tcp*[r]{\textcolor{mintleaf}{\text{Equation 1}}}
%    $Gradient\_update(\mathcal{M}, \_{SFT}(x, CoT)$) \tcp*[r]{\textcolor{mintleaf}{Update model parameters for this batch}}
}
//~\textcolor{mintleaf}{\em Reinforcement learning stage}\;
\For{$i\gets 1$ \KwTo \em $T$}{
	 $\boldsymbol{x},\_ , \boldsymbol{y} \sim  \mathcal{D}_{train}$ \tcp*[r]{\textcolor{mintleaf}{Sample mini-batch without CoT}}
	 $\hat{\boldsymbol{e}} \sim \boldsymbol{\pi}_{\boldsymbol{\theta}} (\boldsymbol{x})$ \tcp*[r]{\textcolor{mintleaf}{On-policy CoT sampling}}
%    $x, \_, y \sim  \mathcal{D}_{train}, \widehat{CoT} \sim \mathcal{M}(x)$\;
    $ \hat{\boldsymbol{y}} \gets \textsc{Extract}(\hat{\boldsymbol{e}})$\tcp*[r]{\textcolor{mintleaf}{Extract the answer from CoT}}
    % $\mathcal{M} \gets \mathcal{M} - lr \nabla L_{rl}(x, y, \widehat{y}, \widehat{CoT})$\;
    % $\boldsymbol{\pi}_{\boldsymbol{\theta}}= \textsc{PPO\_Step}(\boldsymbol{\pi}_{\boldsymbol{\theta}}, \boldsymbol{x}, \boldsymbol{y}, \hat{\boldsymbol{y}}, \hat{\boldsymbol{e}})$ \tcp*[r]{\textcolor{mintleaf}{Perform PPO update}}
    % \textcolor{mintleaf}{// \em PPO update steps}\;
    % $\boldsymbol{\pi}_{\theta_{\boldsymbol{old}}} = \boldsymbol{\pi}_{\boldsymbol{\theta}}$\;
    $\boldsymbol{\pi_{\theta_{\text{old}}}} \gets \boldsymbol{\pi_{\theta}}, V_{\boldsymbol{\phi_{\text{old}}}} \gets V_{\boldsymbol{\phi}}$ \;
    \text{Compute $\delta_t, \hat{A_t}, \hat{R_t}$ using $\pi_{\boldsymbol{\theta_{\text{old}}}}, V_{\boldsymbol{\phi_{\text{old}}}}, \boldsymbol{x}, \hat{\boldsymbol{e}}, \hat{\boldsymbol{y}}$ and $\boldsymbol{y}$}\;% \tcp*[r]{\scriptsize \textcolor{mintleaf}{\S3.1 \textbf{Reinforcement Learning}}}
    % \text{Compute $\delta_t, \hat{A_t}, \hat{r_t}$ using $\boldsymbol{\pi_{\theta_{\text{old}}}} \leftarrow \boldsymbol{\pi_{\theta}}, V_{\boldsymbol{\phi_{\text{old}}}} \leftarrow V_{\boldsymbol{\phi}}$} \;
    \For{$j\gets 1$ \KwTo \em $U$}{
        $\boldsymbol{\theta}, \boldsymbol{\phi} = \textsc{Optimization\_Step}(\mathcal{L}_{RL}(\boldsymbol{\theta}, \boldsymbol{\phi}))$ \tcp*[r]{\textcolor{mintleaf}{Equation 2}}
    }
}
\Return{$\boldsymbol{\pi}_{\boldsymbol{\theta}}$}

\caption{Reinforced Fine-Tuning}
\label{algo:reft}
\end{algorithm*}

% \begin{algorithm}[t!]
% \DontPrintSemicolon
% \KwIn{$\mathcal{D}_{train}: \{(x, r, y)\}$ }
% \myinput{\text{num\_warmup\_step}: $nw$}
% \myinput{\text{num\_rl\_step}: $nr$}
% \myinput{Initial policy: $\mathcal{M}^{init}$}
% \KwOut{Final policy: $\mathcal{M}$}

% $\mathcal{M}=\mathcal{M}^{init}$\;
% \textcolor{mintleaf}{\# \em Warm-up SFT step}\;
% \For{$i\gets 1$ \KwTo $num\_warmup\_step$}{
%     $x, CoT, \_ \sim  \mathcal{D}_{train}$\;
%     % $\mathcal{M} \gets \mathcal{M} - lr \nabla L_{sft}(x, CoT)$\;
%     $Gradient\_update(\mathcal{M}, L_{SFT}(x, CoT)$)
% }
% \textcolor{mintleaf}{\# \em Reinforced learning step}\;
% \For{$i\gets 1$ \KwTo \em $num\_rl\_step$}{
%     $x, \_, y \sim  \mathcal{D}_{train}, \widehat{CoT} \sim \mathcal{M}(x)$\;
%     $ \widehat{y} \gets extract(\widehat{CoT})$\;
%     % $\mathcal{M} \gets \mathcal{M} - lr \nabla L_{rl}(x, y, \widehat{y}, \widehat{CoT})$\;
%     $PPO\_update(\mathcal{M}, x, y, \widehat{y}, \widehat{CoT})$
% }
% \Return{$\mathcal{M}$}\;

% \caption{Reinforced Fine-Tuning}
% \end{algorithm}

\section{Method}
% In this section, we briefly describe the math problem solving task settings and then elaborate Reinforced Fine-Tuning in detail. 
% the general concept can be applied to other tasks.

% \subsection{Task settings}
% In math problem solving, a question is given as input and an answer is expected as output.
% The answer is either a numerical value or single character in the case of MCQ question. \xinbo{Suggestion: These are far from generalizing all math problem solving. It should be pointed out that this is the research scope of datasets in this paper.}The correctness of the predicted answer is based on how well it matches with the golden answer. For numerical value answer, the prediction is correct if and only if it differs from the golden answer by at most $0.01$. For single character answer, the prediction must exactly match the golden answer to be deem correct.\xinbo{suggestion: should these be moved into the metrics part?}

In this work, we focus on \textit{natural language CoT} (\textbf{N-CoT})~\cite{wei2022chain} (Figure \ref{fig:sft_vs_rft}) and \textit{program-based CoT}~\cite{gao2023pal} (\textbf{P-CoT}) using Python. 
\citet{gao2023pal} proposed the program-based CoT for math problem solving. 
We can simply execute the program to obtain the answer. 
To ensure clarity and avoid ambiguity, we use the terms N-CoT and P-CoT to represent natural language and program-based CoTs, respectively.
% in the rest of this paper, respectively.
% N-CoT and P-CoT to represent natural language and program-based CoTs in the rest of this paper.
% ``\textit{rationale}'' to emcompass both CoT and PoT in the rest of this paper.
% }

\subsection{Reinforced Fine-Tuning}
\label{sec:reft}
The proposed Reinforced Fine-Tuning (ReFT) process consists of two stages: the warm-up stage and the reinforcement learning stage. 
The overall algorithm is shown in Algorithm \ref{algo:reft}.
% In WSFT, the model is fine-tuned to generate rationale given a question on a dataset comprising of (question, rationale) tuples. This step teaches the model the basic problem-solving skills to generate a proper response. 
% After that, in RL phase, the model improves its performance via a form of online self-learning using a dataset comprising of (question, answer) tuples.
% Since this step can be done without rationale annotation, a larger dataset can be utilized. 
% However, when the same dataset is used for both phases, ReFT still benefits from exploring and learning from diverse correct rationale.

% \begin{algorithm}[t!]
% \DontPrintSemicolon
% \KwIn{Policy: $\mathcal{M}$}
% \myinput{Question: $x$}
% \myinput{Golden answer: $y$}
% \myinput{Predicted answer: $\widehat{y}$}
% \myinput{Predicted CoT: $\widehat{CoT}$}
% \myinput{PPO epoch: $ppo\_epoch$ default $=4$}
% \KwOut{Updated policy: $\mathcal{M}$}

% \textcolor{mintleaf}{\# \em Update $\mathcal{M}$}\;
% \For{$i\gets 1$ \KwTo $ppo\_epoch$}{
%     \textcolor{mintleaf}{\# \em Compute Reward}\;
%     $\widehat{A}^{GAE(\lambda, \gamma)}$
    
%     \textcolor{mintleaf}{\# \em Compute Advantage}\;
%     $\widehat{A}^{GAE(\lambda, \gamma)}$

% }

% \Return{$\mathcal{M}$}\;

% \caption{PPO update}
% \end{algorithm}

% The same dataset can be used to train models in both phases.
% Both phase can be done using the same set of data.

\paragraph{Warm-up}
% In WSFT, the model is fine-tuned on a dataset comprising of (question, rationale) tuples. This step teaches the model the basic problem-solving skills to generate a proper response. Language modelling objective is used for fine-tuning and the loss is computed from all tokens in the rationale. 
% In this phase, the model learns the basic problem-solving skills to generate a proper response from a dataset comprising of (question, rationale) tuples. 
% In the two examples shown in [], the rationale can be natural language description formatted in a way that the answer can be easily extracted or it could be a program whose executed result is the answer.
% In this phase, the model is fine-tuned using standard language modelling objective in which the loss is computed from all rationale tokens. 
% More specifically, given a dataset $\mathcal{D}$ consisting of question $x$ and rationale $CoT$ tuples, the loss function is defined in equation \ref{eq:sft-loss} where $CoT$ is represented as a sequence of tokens $\{CoT_i\}_{i=1}^{L}$.
% \begin{equation}
%     \mathcal{L}_{SFT}=-\mathbb{E}_{(x,CoT)\sim \mathcal{D}}\left[\sum_{i=1}^{L} \text{log}\left[P\left(CoT_i|x, CoT_{<i}\right)\right]\right]
%     \label{eq:sft-loss}
% \end{equation}
% Given a dataset $\mathcal{D}$ consisting of question $x$ and rationale $R$ tuples, let's denote the state ${s_t}$ as a tuple of question and the rationale tokens up to and including its $i^{th}$ tokens $(x, CoT_{:i})$ and $a_t$ as any
% The policy models the probablity of the next tokens $\pi(a_t|s_t)$ where $a_t$ can be any vocabulary token.
In this stage, the policy is fine-tuned for a few epochs on a dataset comprising of the ``(\textit{question}, \textit{CoT})'' tuples: $(\boldsymbol{x}, \boldsymbol{e})$.
It enables the model to have basic problem-solving skills to generate a proper response\footnote{The underlying concept is similar to the verifier training~\cite{cobbe2021training} to generate multiple solutions.}. 
Formally, the CoT generation process can be decomposed into a sequence of next token prediction actions. 
The last action token, \texttt{<eos>}, signals the generation process to terminate. 
The CoT $\boldsymbol{e}$ is written as:
% $$
%     \textit{rationale} = [a_1, a_2, ..., a_{L-1}, a_L\texttt{=<eos>}]
% $$
$$
    \boldsymbol{e} = [a_1, a_2, ..., a_{L-1}, a_L\texttt{=<eos>}]
$$
where $L$ represents the maximum length. 
At timestep $t$, the action $a_t$ is sampled from a policy $\boldsymbol{\pi}_{\boldsymbol{\theta}}(\cdot|s_t)$ where $a_t$ can be any token in the vocabulary and the state $s_t$ comprises of all tokens in the question and all tokens generated so far. 
After each action, the resulting state $s_{t+1}$ is the concatenation of the current state $s_t$ and the action $a_t$:
% $$
%     s_{t+1} = [s_t, a_t]
% $$
$$
    s_{t+1} = 
        \begin{cases}
            \boldsymbol{x}, & t = 0 \\
            [s_t, a_t],     & 1 \leq t \leq L \\
        \end{cases}.
$$
% \allan{Is it better to use Uppercase $S_t$ as ``state''??? Feel free to remove this if feels fine about lowercase} -- Peng: lower case state s_t is actually standard in the literature, I think it's fine:)
% As the produced action corresponds to the \texttt{<eos>} token, 
As the produced action is the \texttt{<eos>} token, 
the resulting state $s_{L+1}$ is the terminal state and the generation process is finished.
With this notation, the loss function for a sample can be written as:
\begin{equation}
    \mathcal{L}_{SFT}(\boldsymbol{\theta})=-\mathbb{E}_{\boldsymbol{e} \sim \mathcal{D}}\left[\sum_{t=1}^{L}\log\left(\boldsymbol{\pi}_{\boldsymbol{\theta}}(a_t\vert s_t)\right)\right]
    \label{eq:sft-loss}
\end{equation}

% \peng{Suggestion: define early in this place the state $s_t$, action $a_t$, policy $\pi(a_t|s_t)$, and rewrite Eq.~\ref{eq:sft-loss} with the notations.}
% In the case that the same dataset is used for both phases, it is important not to let the model overfit on this data during WSFT. Otherwise, in RL phase, the model will fail to explore any new  correct rationales as it is stuck in the local optima. To overcome this, we limit that WSFT is done for just few step/epoch so that the training set accuracy is within a reasonable range (e.g between 50\%-80\%)
% In the case that the same dataset is used for both phases, the model should not overfit on this data during WSFT because good sampling diversity is required to make RL training phase feasible. To overcome this, WSFT is to be done for a short duration so that the training performance is within a reasonable range (e.g between 50\%-80\%)

\paragraph{Reinforcement Learning}
In this stage, the policy improves its performance via a form of online self-learning using a dataset comprising of (\textit{question}, \textit{answer}) tuples: $(\boldsymbol{x}, \boldsymbol{y})$. 
Specifically, the policy model learns by repeatedly sampling responses (Figure \ref{fig:sft_vs_ref_path}), evaluating the response's answer correctness, and updating its parameters in an online fashion (line 7-14 in Algorithm \ref{algo:reft}).
We employ PPO~\cite{schulman2017proximal} with a clipped objective algorithm for training. 
Following \citet{ziegler2019fine}, the value model $V_{\phi}$ is constructed by appending a linear value head on top of the last hidden states of the policy model $\pi_{\theta}$, which is the model after the warm-up stage.
% Since this step can be done without rationale annotation, a larger dataset can be utilized. 
% However, when the same dataset is used for both phases, ReFT still benefits from exploring and learning from diverse correct rationale.
% In this phase, the model learns by repeatedly sample response, evaluate the response's answer correctness and update its parameters in an online fashion. We employ PPO with clipped objective algorithm for training. More specifically, following [], the value model $V_{\phi}$ is constructed by appending a linear value head on top of the last hidden states of the policy model $\pi_{\theta}$, which is the model after WSFT phase.
The reward of 0 is given for all action resulting in non-terminal state. 
At the terminal state, we use a reward function that directly compares the answer extracted from the state's CoT and the ground-truth answer $\boldsymbol{y}$ . 
Here, the reward function returns 1 if the answer is deemed correct, otherwise 0 is returned. 
On dataset whose answers are all numeric, \textit{partial reward}~\cite{zhong2017seq2sql,le2022coderl} of 0.1 can be applied when the answer can be extracted and it is of numeric type. 
For $1 \leq t \leq L$, we write
% Such partial reward setting has also been proven to be effective in reinforcement learning~\cite{zhong2017seq2sql,le2022coderl}.
% \begin{equation}
%     r(\boldsymbol{x},\hat{y}) = 
%         \begin{cases} 
%             1, &  \hat{y} = y^{*} \\
%             0.1, & \hat{y} \neq \texttt{null} \\
%             0, & \hat{y} = \texttt{null}
%     \end{cases}
%     \label{equ:reward_function}
% \end{equation}
\begin{equation}
    r(s_t, a_t, s_{t+1})\! =\! 
        \begin{cases} 
            1, &  \!\!\!\!\! \texttt{EXTRACT}(s_{t+1}) = \boldsymbol{y} \\
            0.1, &\!\!\!\!\! \texttt{EXTRACT}(s_{t+1}) \neq \texttt{null}, \neq \boldsymbol{y} \\
            0, & \!\!\!\!\! \texttt{EXTRACT}(s_{t+1}) = \texttt{null}
    \end{cases}
    \nonumber
    \label{equ:reward_function}
\end{equation}
Such a partial reward can help reduce the effect of learning from sparse reward~\cite{riedmiller2018learning,trott2019keeping}.
% \peng{Write explicitly the immediate reward $r_t := r(s_t,a_t)$}
In addition, following \citet{zheng2023secrets}, our total reward is the sum of the reward function score and the Kullback-Leibler (KL) divergence~\cite{kullback1951information} between the learned RL policy and initial policy scaled by a coefficient factor $\beta$. 
\begin{equation*}
    \begin{split}
        r_{total}(s_{t}, & a_{t},s_{t+1})=r(s_{t},a_{t},s_{t+1}) \\ & - \beta \textit{KL}\left(\boldsymbol{\pi}_{\boldsymbol{\theta}}(\cdot|s_t), \boldsymbol{\pi}_{\boldsymbol{\theta}}^{(0)}(\cdot|s_t)\right)
    \end{split}
\end{equation*}
The generalized advantage estimate~\cite{schulman2018highdimensional} is used for advantage calculation:
% We employ the generalized advantage estimate~\cite{schulman2018highdimensional} for advantage calculation:
% For advantage calculation, the generalized advantage estimate from \citet{schulman2018highdimensional} is employed.
\begin{equation*}\label{eq:Advantage}
    \hat{A_t} = \sum_{l=0}^{L-t} (\gamma\lambda)^{l}\delta_{t+l},
\end{equation*}
%where $k$ denotes the number of remaining valid tokens from position $t$ to the end of this sequence, 
where the Temporal Difference (TD) is defined as
\begin{equation*}
    \delta_{t'} = -V_{\phi}(s_{t'}) + r_{total}(s_{t'},a_{t'},s_{t'+1}) + \gamma V_{\phi}(s_{t'+1})
\end{equation*}
with the terminal state value $V_{\phi}(s_{L+1}) := 0$,
$\lambda \in (0,1]$ is the discount factor for rewards,
and $\gamma \in [0,1]$ is the discount factor for TD.
For the estimate of return, 
we leverages the $\lambda$-return $\hat{R_t}$, 
which can be written as the sum of the generalized advantage estimate and the value estimate:
\begin{equation*}
    \hat{R_t} = \hat{A_t} + V_{\boldsymbol{\phi}}(s_t)
\end{equation*}
% Finally, the loss function can be written as in Equation~\ref{eq:loss_rl}.
Lastly, the policy and value objectives can be written as in two equations below
% Equation~\ref{eq:loss_policy} and Equation~\ref{eq:loss_value}
% \begin{multline}\label{eq:loss_rl}
%     L_{RL}(\theta) = -\mathbb{E} \left[ \text{min}\left(\dfrac{\pi_{\theta}\left(a_t|s_t\right)}{\pi_{\theta_{old}}\left(a_t|s_t\right)}\hat{A_t}, \\ \text{clip}\left(\dfrac{\pi_{\theta}\left(a_t|s_t\right)}{\pi_{\theta_{old}}\left(a_t|s_t\right)}, 1-\epsilon, 1+\epsilon\right)\hat{A_t}\right) \right]
% \end{multline}
\begin{equation*}
\begin{split}
% \mathcal{L}_{RL}(&\boldsymbol{\theta})  = -\mathbb{E}\Bigg[ \min \Bigg( \frac{\boldsymbol{\pi}_{\boldsymbol{\theta}} (a_t|s_t)}{\boldsymbol{\pi}_{\text{old}} (a_t|s_t)} \hat{A}_t, \\
% {\boldsymbol{\pi}_{\theta_{\text{old}}}}
\mathcal{L}_{policy}(&\boldsymbol{\theta})  = -\mathbb{E_{\boldsymbol{e} \sim \boldsymbol{\pi}_{\theta_{\text{old}}}}}\Bigg[ \min \Bigg( \frac{\boldsymbol{\pi}_{\boldsymbol{\theta}} (a_t|s_t)}{\boldsymbol{\pi}_{\theta_{\text{old}}} (a_t|s_t)} \hat{A}_t, \\
&\text{clip}\left( \frac{\boldsymbol{\pi}_{\boldsymbol{\theta}} (a_t|s_t)}{\boldsymbol{\pi}_{\theta_{\text{old}}} (a_t|s_t)}, 1-\epsilon, 1+\epsilon \right) \hat{A}_t \Bigg) \Bigg]
\end{split}
% \label{eq:loss_rl}
\label{eq:loss_policy}
\end{equation*}
% \begin{multline}
%     L_{RL}(\theta) = -\mathbb{E} \left[ \text{min}\left(\rho_t(\theta)\hat{A_t}, \\ \text{clip}\left(\rho_t(\theta), 1-\epsilon, 1+\epsilon\right)\hat{A_t}\right) \right]
% \end{multline}
% \begin{multline}
%     \rho_t(\theta) = \dfrac{\pi_{\theta}\left(a_t|s_t\right)}{\pi_{\theta_{old}}\left(a_t|s_t\right)}
% \end{multline}
% where $\boldsymbol{\pi}_{\theta_{\text{old}}}$ is the policy used for sampling CoT,
\begin{equation*}
\begin{split}
\mathcal{L}_{value}(&\boldsymbol{\phi})  = \frac{1}{2}~ \mathbb{E_{\boldsymbol{e} \sim \boldsymbol{\pi}_{\theta_{\text{old}}}}}\Bigg[
\!\! \max \Bigg(\!\! \norm{V_{\boldsymbol{\phi}}(s_t) - \hat{R_t}}^2, \\ 
&\norm{\text{clip}\left(\hat{R_t}-V_{\phi}(s_t), \hat{A_t} -\epsilon, \hat{A_t} + \epsilon \right)}^2 \Bigg) \Bigg]
\end{split}
\label{eq:loss_value}
\end{equation*}
where $\boldsymbol{\pi}_{\theta_{\text{old}}}$, $V_{\phi_{\text{old}}}$ are used for sampling CoT and computing $\hat{A_t}$, $\hat{R_t}$.
% the return $\hat{r_t}$ is the sum of the advantage and the value.
% \begin{equation*}
%     \hat{r_t} = \hat{A_t} + V_{\phi_{old}}(s_t)
% \end{equation*}
The unified loss function is the weighted sum of the above objectives.
\begin{equation}\label{eq:loss_rl}
    \mathcal{L}_{RL}(\boldsymbol{\theta}, \boldsymbol{\phi}) = \mathcal{L}_{policy} + \alpha \mathcal{L}_{value}
\end{equation}
where $\alpha$ is the coefficient for the value objective.
% Refer to Appendix \ref{sec:appendix_ppo} for the PPO training details.
% Refer to algorithm [] for the code snippet of the update step.
% \peng{use $\rho$ to denote the ratio and avoid the notational conflict.} -- modifications done.

\section{Experiments}

% To demonstrate that ReFT leads to superior policy, 
% We conduct experiments to compare ReFT with three baselines: SFT and two variants of self-training methods on three benchmark datasets: GSM8K, MathQA and SVAMP.  
% We use Galactica-6.7B\footnote{\url{https://huggingface.co/facebook/galactica-6.7b}} and Codellama-7B\footnote{\url{https://huggingface.co/codellama/CodeLlama-7b-hf}} as foundation models and employ both natural language rationale (CoT) and program based rationale (PoT). The policy top-1 accuracy is used as metric for the comparison.

% In addition, we also apply common techniques: majority voting and reward model reranking on ReFT policy to verify if they are still beneficial. We choose GSM8k as the benchmark, report voting@100, rerank@100 performances and compare with other works.

% Reward Hacking ...

% Lastly, to investigate if ReFT is robust to model scale, we also experiment with using Galactica-125m as foundation model.

% and three evaluation metrics: greedy generation accuracy (\textbf{Acc}), majority voting accuracy (\textbf{Voting@100}) and re-ranking with reward model accuracy (\textbf{Rerank@100})

\subsection{Datasets}
We conduct experiments on three math problem datasets: GSM8K~\cite{cobbe2021training}, SVAMP~\cite{patel2021nlp} and MathQA~\cite{amini2019mathqa}.
For both GSM8K and SVAMP, the format of answers is a numeric value.
In MathQA, the format is instead a list of multiple choices (i.e., \texttt{ABCD}).
Table \ref{tab:data_statistics} presents the statistics of all datasets.
We perform few-shot prompting~\cite{wei2022chain,gao2023pal} using \texttt{GPT-3.5-turbo} to obtain both the N-CoT and P-CoT annotations\footnote{Examples of N-CoT and P-CoT representations can be found in Appendix \ref{sec:appendix_representation}.}.
% We perform few-shot prompting~\cite{wei2022chain,gao2023pal} using \texttt{GPT-3.5-turbo} to obtain both the N-CoT and P-CoT annotations\footnote{See Appendix \ref{sec:appendix_representation} for example of N-CoT and P-CoT.}.
% \allan{I removed this citation~\cite{jie2023design}, just not sure if it is appropriate to cite our own article here. Feel free to add it back and remove this comment.} -- Peng: I think it's okay to add it back, as we indeed following the settings of jie2023design for the desing of CoT and PoT...
The N-CoT and P-CoT annotations are obtained following \citet{jie2023design}.
% The training data of GSM8k, SVAMP and MathQA datasets are obtained following ~\cite{jie2023design}. 
We also conducted an additional experiment on a numeric version of MathQA~\cite{jie2023leveraging} where the format is also a numeric value.
Such experiments are used to demonstrate our assumptions of potential reward hacking phenomenon~\cite{skalse2022defining} on MathQA (\S\ref{sec:results}).

% The MathQA$_\text{numeric}$ numeric benchmark is obtained from ~\cite{jie2022learning}. 
% We present the data statistics in table \ref{tab:data_statistics}.

\begin{table}[t!]
    \centering
    \adjustbox{max width=1\linewidth}{
    \begin{tabular}{ccccc}
    \toprule
        \textbf{} & \textbf{GSM8k} & \textbf{SVAMP} & \textbf{MathQA$_\text{MCQ}$} & \textbf{MathQA$_\text{numeric}$} \\ \midrule
    \textbf{Train N-CoT}  & 7,465  & 3,076  & 14,862  & 8,955           \\ 
    \textbf{Train P-CoT}  & 7,356  & 3,043  & 15,250  & 7,672           \\ 
    \textbf{Test} & 1,319  & 1,000  & \textcolor{white}{0}1,605   & 1,605          \\ 
    \bottomrule
    \end{tabular}
    }
    % \vspace*{-3mm}
% \caption{Dataset statistics of two types of CoT in the training set and the test set.}
\caption{Statistics of the train and test datasets.}
\label{tab:data_statistics}
\end{table}

\subsection{Baseline}
\label{sec:baseline}
% Supervised Fine-Tuning and Self-Training.
% PPO with positive samples only.
We compare ReFT with SFT and self-training~\cite{xie2020self,amini2022self} baselines.
SFT simply fine-tunes the language model on the training data.
% Experiments with self-training methods ensure a relatively fair comparison because all these methods share the mechanism that the training makes use of the samples generated from the model.
Experiments with self-training methods ensure a relatively fair comparison because these methods share the mechanism that the samples generated from the model are used for training.

% The SFT process directly fine-tunes the language model for 40 epochs\footnote{This number has been chosen sufficiently large to ensure the model converges.} and chooses the checkpoint with the best accuracy. 
% and two variants of self-training methods.
% Three baseline methods are considered in this work are Supervised Fine Tuning (SFT), Offline Self-Learning (Offline-SL) and Online Self-Learning (Online-SL). 
% In SFT method, the baseline is obtained by finetuning a foundation model on the training data.

We implemented Offline Self-Training (\textbf{Offline-ST})~\cite{he2020revisiting}, and Online~\cite{hoi2021online} Self-Training (\textbf{Online-ST}).
The Offline-ST method is similar to expert iteration~\cite{anthony2017thinking,uesato2022solving,zelikman2022star}.
We first use the SFT checkpoint from the early checkpoint to sample the CoTs and verify them against the ground truth. 
% The temperature for sampling is set to $1.0$ and the max length is $1024$. 
We only retain those expert samples that have a correct answer.
We perform SFT on the combination of original training data and the expert samples.
% \footnote{The performance is a little lower if only expert samples are used}.
% We perform SFT on the combination of original training data and the expert samples.
% \footnote{The performance is lower if only expert samples are used}.

% In Offline-SL method, we first use one of SFT model checkpoint to sample rationales, and keep only correct ones. 
% Then, the Offline-SL baseline is obtained by finetuning a foundation model on the combination of data with collected rationales and original training data.
The Online-ST method is made to be closely comparable to ReFT.
Following ReFT, Online-ST has the same warm-up process.
After that, we perform continual training with the samples generated on the fly.
At each training step, the model first samples CoTs for a batch and only retains those with correct answers. 
The resulting batch consists of both sampled and ground-truth CoTs.
% \footnote{The inclusion of ground-truth CoTs is necessary to stabilize the training and prevent performance collapse}. 
We then update the model parameters on this batch with the supervised fine-tuning objective $\mathcal{L}_{SFT}$.
% both sampled rationale and ground-truth rationales
% Online-ST initializes the model from an WSFT checkpoint, then the training is done online as the model parameter is updated while looping through the training data. 
% At each training step, the model first samples rationales for a batch and only retains those with correct answers. 
% Then, the combined batch of collected rationales and original rationales is used to update the model parameter with $\mathcal{L}_{SFT}$.
Compared with ReFT, Online-ST neither makes use of negative responses (with an incorrect answer) nor has a dedicated mechanism to prevent the model from significantly diverging from the initial model, which can manifest as task-specific overfitting and training instability.

\begin{table*}[t!]
    \centering
    \adjustbox{max width=1.0\linewidth}{
    \begin{tabular}{lccccccccc}
        \toprule
        \multirow{2}{*}{ \bf Method } & \multirow{2}{*}{ \bf Size } & \multicolumn{2}{c}{\bf GSM8K} & \multicolumn{2}{c}{\bf SVAMP} & \multicolumn{2}{c}{\textbf{MathQA}$_\text{MCQ}$} & \multicolumn{2}{c}{\bf Average}\\
        &  & \textbf{N-CoT} & \textbf{P-CoT} & \textbf{N-CoT} & \textbf{P-CoT} & \textbf{N-CoT} & \textbf{P-CoT} & \textbf{N-CoT} & \textbf{P-CoT}  \\
        \midrule
Galactica + SFT & 6.7B                                      & $42.68$ & $58.83$ & $54.50$ & $70.09$ & $58.07$ & $64.61$ 	& $51.75$ & $64.51$ \\
Galactica + Offline Self-Training & 6.7B                    & $42.60$ & $60.72$ & $57.90$ & $72.30$ & $\mathbf{60.75}$ & $67.04$ 	& $53.75$ & $66.69$ \\
Galactica + Online Self-Training  &6.7B                     & $47.84$ & $62.93$ & $59.40$ & $\mathbf{74.59}$ & $59.38$ & $61.24$ 	& $55.54$ & $66.25$ \\
Galactica + ReFT &  6.7B                                    & $\mathbf{48.14}$ & $\mathbf{68.91}$ & $\mathbf{61.40}$ & $74.09$ & $58.13$ & $\mathbf{70.47}$ 	& $\mathbf{55.89}$ & $\mathbf{71.16}$ \\
        \midrule
        \midrule
CodeLLAMA + SFT & \textcolor{white}{0.}7B                   & $43.59$ & $63.68$ & $58.09$ & $75.40$ & $56.01$ & $64.79$ 	& $52.56$ & $67.96$ \\
CodeLLAMA + Offline Self-Training & \textcolor{white}{0.}7B & $45.10$ & $68.00$ & $60.20$ & $77.69$ & $59.81$ & $68.53$ 	& $55.04$ & $71.41$ \\
CodeLLAMA + Online Self-Training &\textcolor{white}{0.}7B   & $44.66$ & $67.85$ & $58.60$ & $77.40$ & $56.95$ & $68.85$ 	& $53.40$ & $71.37$ \\
CodeLLAMA + ReFT & \textcolor{white}{0.}7B                  & $\mathbf{53.30}$ & $\mathbf{75.28}$ & $\mathbf{64.50}$ & $\mathbf{79.19}$ & $\mathbf{60.13}$ & $\mathbf{71.83}$ 	& $\mathbf{59.31}$ & $\mathbf{75.43}$ \\
        \bottomrule
    \end{tabular}
    }
    % \vspace*{-2.5mm}
    % \caption{Value accuracy comparison among the baselines and proposed ReFT method fine-tuned with two foundation models on all datasets.
    \caption{Value accuracy of ReFT and the baselines fine-tuned with two foundation models on all datasets.
    % \xinbo{do we need to put the results of MathQA$_\text{numeric}$ here?}
    }
    \label{tab:sft_ppo_result}
\end{table*}

\subsection{Experimental Setup}
\label{sec:setup}

% We conduct experiments with two foundation models: Galactica-6.7B\footnote{\url{https://huggingface.co/facebook/galactica-6.7b}}~\cite{taylor2022galactica} and CodeLLAMA-7B\footnote{\url{https://huggingface.co/codellama/CodeLlama-7b-hf}}~\cite{roziere2023code}. 
We conduct experiments with two foundation models: Galactica-6.7B\footnote{\rurl{huggingface.co/facebook/galactica-6.7b}}~\cite{taylor2022galactica} and CodeLLAMA-7B\footnote{\rurl{huggingface.co/codellama/CodeLlama-7b-hf}}\footnote{Additional preliminary experiments were conducted using Gemma~\cite{gemmateam2024gemma}. However, these results are not included in the current version of this paper due to unresolved implementation issues that align with known challenges reported within the open-source community (\url{https://huggingface.co/google/gemma-7b/discussions}).}~\cite{roziere2023code}.
% Both models are reported to have strong performance in solving math problems and are commonly adopted in recent literature on reasoning 
Both models are reported to have strong performance in math solving and are commonly adopted in recent literature on reasoning tasks~\cite{yue2023mammoth,luo2023wizardmath}.

In addition to the comparison with baselines, we also apply common techniques, majority voting~\cite{wang2022self} and reward model reranking~\cite{lightman2023lets} on GSM8K. 
% We use Galactica-6.7B\footnote{\url{https://huggingface.co/facebook/galactica-6.7b}} and Codellama-7B\footnote{\url{https://huggingface.co/codellama/CodeLlama-7b-hf}} as foundation models and employ both natural language rationale (CoT) and program based rationale (PoT). The policy top-1 accuracy is used as metric for the comparison.

% In addition, we also apply common techniques: majority voting and reward model reranking on ReFT policy to verify if they are still beneficial. We choose GSM8k as the benchmark, report voting@100, rerank@100 performances and compare with other works.

\paragraph{Hyper-parameters}
In all experiments, the training is done with 8 A100-80GB GPUs using DeepSpeed~\cite{rajbhandari2020zero,rasley2020deepspeed} Zero stage 2 and HuggingFace Accelerate~\cite{accelerate}.
During the warm-up stage of ReFT, we use AdamW~\cite{loshchilov2017decoupled} optimizer with 10\% warm-up ratio. 
% The batch size is set to 48 and learning rate is $1e$-$5$.
The batch size is 48 and learning rate is $1e$-$5$.
 % using AdamW~\cite{loshchilov2017decoupled} optimizer with a learning rate of  $3e$-$6$ and 0.1 warm-up ratio.
 % and the learning rate schedule that linearly warm-up from 0 to $3e-6$ after first 10\% of the total optimization steps then linearly decay to zero. 
The maximum length is set to $1024$. 
% The maximum length of both question and response is set to $1024$. 
% The number of epoch is chosen such that the model performance on the training set is between 50\%-80\%.
% With CodeLlama foundation model, the WSFT is done for 1 epoch for SVAMP and 2 epochs for GSM8k and MathQA.
The number of epochs in the warm-up stage is 2 in all settings except on MathQA$_\text{MCQ}$ and MathQA$_\text{numeric}$ where we use up to 5 and 10 respectively. 
% During reinforcement learning, the global batch size is set to 32 and the maximum length is set to $700$ for efficient training.
The model is trained for $300$ epochs with a learning rate of $3e$-$7$.
Following \citet{ziegler2019fine}, the $\lambda$, $\gamma$, $\alpha$, $\epsilon$ and $U$ in PPO are set to $1$, $0.95$, $5$, $0.2$, and $2$, respectively.
% The hyper-parameter learning rate, $\lambda$, $\gamma$ and $\epsilon$ are $3e-7$, $1$, $0.95$, and $0.2$ respectively. 
The KL coefficient $\beta$ is set to $0.01$ for P-CoT and is set to $0.05$ for N-CoT experiments. 
Further hyperprameter settings about ReFT can be found in Appendix \ref{sec:appendix_hyperparameter}.

% \paragraph{Baselines}
% For SFT method, we follow WSFT settings except that the model is fine-tuned for 40 epochs. The number of epoch is set to be sufficiently large to capture the upper-bound SFT performance. We choose the checkpoint with highest top-1 accuracy on the test set.
For SFT baseline, we train the model for 40 epochs and choose the checkpoint with best performance. 
% For SFT method, we directly choose the checkpoint with the highest testing set top-1 accuracy from 40 checkpoints produced from WSFT phase. 
This number of epochs has been chosen to be sufficiently large to ensure SFT converges.
% capture the SFT upper-bound performance.
For Offline-ST baseline, we sample the CoTs by using the checkpoint from the ReFT warm-up stage.
% ensure good sampling diversity. 
% model from WSFT is used to sample responses to ensure good sampling diversity. 
Using the generation temperature of 1.0 and max length of 1024, we sample 100 CoTs for each question and only keep those with a correct answer. 
Following \citet{singh2023human}, we then sub-sample the CoTs to 10 random unique CoTs per question to balance difficulties of questions.
The number of fine-tune epoch is set to 20, which is sufficiently large to ensure the training to converge.
% the number of data between easy and hard questions. 
% The same settings in SFT method is used for both training and evaluation.
As mentioned in \S\ref{sec:baseline}, the Online-ST baseline tries to mimic the same setting as in ReFT. 
We have the same warm-up process and the hyperparameter setting is roughly the same as ReFT.
% we initialize model from WSFT and follow RL stage setting as close as possible. e.g  learning rate, batch size, number of model updates for every batch, e.t.c.

\paragraph{Reward Model Reranking}
\label{sec:reward_model_reranking}
Following \cite{cobbe2021training,uesato2022solving}, we train a reward model (RM) to determine the correctness of the CoT. 
% a reward model (RM) is trained to determine whether an answer to the question is correct or not. 
To construct the RM training data, we use the model from the warm-up stage and perform sampling to obtain 100 CoTs for each question in the training set.
% To construct the RM training data, we use the model from WSFT and perform sampling to obtain a 100 rationales for each question in the SFT dataset.
The CoTs are deduplicated and the binary labels can be obtained by comparing the extracted answer against the ground truth.

As a common practice, the reward model is a language model that is initialized from the best SFT checkpoint~\cite{cobbe2021training,ouyang2022training}. 
Similar to the outcome-based reward model (ORM) \cite{uesato2022solving}, the reward model is trained to predict a binary label that indicates the ``\textit{correct}'' or ``\textit{incorrect}'' solution. Once the input passes through the reward model, classification is conducted with a linear classifier on the hidden state of the last token. 
Finally, the solution with the highest ``correct'' score among the candidates is selected as the final answer. 
We train the RM model for 3 epochs using a batch size of 24, the maximum length of 700 and a linear learning rate schedule with $10$\% warm-up period and the max learning rate of $1$e$-6$.

\paragraph{Evaluation}
We report value accuracy for both N-CoT and P-CoT on all datasets. 
% Specifically for majority voting and reranking (Table \ref{tab:voting_reranking}), we sample 100 CoTs for evaluation. 
For majority voting and reranking (Table \ref{tab:voting_reranking}), we sample 100 CoTs for evaluation. 
In voting, the valid answer with majority counts is chosen as the final answer for computing accuracy.
In reranking, we choose the CoT with the highest score and extract the answer. 
% \textbf{Top-1 Accuracy}: The model generates rationale for a question using greedy search. The extracted answer is used for computing accuracy.\\
% \textbf{Voting@100}: The model samples 100 rationales and extracts the answers. The valid answer with majority counts is chosen as the final answer for computing accuracy.\\
% \textbf{Rerank@100}: The model samples 100 rationales and extracts the answers. 
% Then, a reward model is used to score the rationales that produce a valid answer (i.e., not \texttt{null}). 
% The answer from the best rationale is chosen as the final answer for computing accuracy.
% Detail of reward model can be found in section \ref{sec:reward_model_reranking}.

\subsection{Results}
\label{sec:results}

\paragraph{ReFT Outperforms SFT}
Table \ref{tab:sft_ppo_result} compares the performance among the baselines and proposed ReFT on GSM8K, SVAMP, and MathQA datasets. 
We can observe that ReFT consistently achieves much better performance over the SFT 
% and the self-training family approaches 
except on MathQA$_\text{MCQ}$ N-CoT.
Specifically, we have closed to $10$-point and $12$-point improvement over SFT with CodeLLAMA on GSM8K N-CoT and P-CoT, respectively. 
On average, we achieve $6.7$-point and $7.4$-point improvements with CodeLLAMA on all datasets in N-CoT and P-CoT, respectively. 
% More importantly, no additional annotations or reward models are used in ReFT. 
Notably, no additional annotations or reward models are used in ReFT. 
Such strong results demonstrate robust generalization of ReFT (see Analysis \S\ref{sec:analysis}) and huge potential for further exploring the training data with reinforcement learning~\cite{lu2023reinforcement}. 

Offline self-training includes the sampling data from the initial policy for fine-tuning. 
We can see this simple baseline can improve the performance compared with SFT~\cite{he2020revisiting,gulcehre2023reinforced} but the improvements are far behind the one made by ReFT. 
Such comparisons indicate that ``\textit{exploring}'' is essential in ReFT to have good performance. 
Though online self-training achieves some more improvements with Galactica, 
% its training is sometimes unstable\footnote{We sometimes noticed the performance drops significantly and unable to recover} and the performance 
it is still far behind ReFT on average.
This result indicates that incorrect instances are also very essential to guide the model for better exploration. 
Comparisons with self-training also suggest the proposed approach with on-policy sampling and reinforcement learning is better than standard data augmentation approaches. 
% Both online and offline self-training adopt the sampling data from the policy. 
% We can observe certain improvements by self-training. 
% However, ReFT still outperforms these methods on average, indicating better exploration by ReFT 

\begin{figure}[t!]
    \centering
    \adjustbox{max width=1.0\linewidth}{
		\begin{tikzpicture}[node distance=2.0mm and 2.0mm, >=Stealth, 
			wordnode/.style={draw=none, minimum height=5mm, inner sep=0pt},
			chainLine/.style={line width=0.8pt,-, color=mygray},
			entbox/.style={draw=black, rounded corners, fill=red!20, dashed},
			mathop/.style={draw=none, circle, minimum height=2mm, inner sep=1pt, line width=0.8pt, fill=pinkgla},
			stepsty/.style={draw=pinkgla, circle, minimum height=2mm, inner sep=1pt, line width=0.8pt},
			quant/.style={draw=none, minimum height=2mm, inner sep=1pt, line width=1pt},
			expr/.style={draw=electronblue, rectangle, minimum width=20mm, minimum height=4.5mm, inner sep=2pt, line width=0.8pt, rounded corners},
			box/.style={draw=electronblue, rectangle, rounded corners, line width=0.9pt, dashed, minimum width=26mm, minimum height=20mm},
			invis/.style={draw=none},
            highlight/.style={fill=yellow!50, rounded corners}
			]

            \node[wordnode, align=left, text width=12cm, font=\fontfamily{phv}\selectfont] (question) [] {\textbf{Question}: The diagonals of a rhombus are 18 cm and 22 cm. Find its area?};

            \node[wordnode, below=of question, align=left, font=\fontfamily{phv}\selectfont, text width=12cm] (options) [] {A) 277,  B) 266, \textcolor{white}{{C) 198}}, D) 288, E) 212};
			
			% Separate node for the correct option C
            \node[wordnode, align=left, font=\fontfamily{phv}\selectfont] (correctOption) at (options.south west) [xshift=3.36cm, yshift=0.22cm] {\underline{\bf C) 198}};

            \node[wordnode, align=left, text width=12cm, font=\fontfamily{phv}\selectfont, below=of options, draw=electronblue, inner sep=3pt, line width=1.5pt] (cot) [] {\textbf{Generated CoT}: To find the area of a rhombus, we need to multiply the lengths of its diagonals and divide the result by 2. \\
            Area of rhombus = (Product of diagonals) / 2 \\
            Area of rhombus = (18 cm x 22 cm) / 2 \\
            Area of rhombus = 344 cm$^2$ / 2 \\
            Area of rhombus = \textcolor{red}{172 cm$^2$} \\
            Therefore, the answer is: C };
                        
            % Highlight box for the correct option
            \begin{pgfonlayer}{background}
				\node[highlight, fit=(correctOption)] {};
			\end{pgfonlayer}

		\end{tikzpicture}
	}
	\vspace*{-5mm}
    \caption{Example prediction of MathQA$_\text{MCQ}$ reveals reward hacking.}
    \label{fig:reward_hacking}
\end{figure}

\begin{table}[t!]
    \centering
    \adjustbox{max width=1.0\linewidth}{
        \begin{tabular}{llc}
        \toprule
        \textbf{N-CoT} & \textbf{Galactica} &  \textbf{CodeLLAMA}   
        \\ \midrule
        SFT   & $40.08$   &  $37.32$ \\ 
        Offline Seft-Training & $44.23$   &  $41.24$ \\ 
        Online Seft-Training  & $43.78$   &  $38.06$ \\ 
        ReFT  & $\mathbf{45.23}$ &  $\mathbf{42.24}$  \\ 
        \bottomrule
        \end{tabular}
    }
    % \vspace*{-3mm}
\caption{Value accuracy of ReFT and the baselines with two foundation models on MathQA$_\text{numeric}$ benchmark}
\label{tab:mathqa_numeric}
\end{table}

% \begin{table}[t!]
%     \centering
%     \adjustbox{max width=0.5\linewidth}{
%         \begin{tabular}{llc}
%         \toprule
%         \multicolumn{2}{r}{\bf Method}  & \textbf{N-CoT}         \\
%         \midrule
%         \multirow{2}{*}{\textbf{Galactica}} & SFT  & $40.08$     \\ 
%                                             & ReFT & $45.23$     \\ 
%         \midrule
%         \multirow{2}{*}{\textbf{CodeLLAMA}} & SFT   & $37.32$    \\ 
%                                             & ReFT  & $42.24$    \\ 
%         \bottomrule
%         \end{tabular}
%     }
% \caption{Accuracy of SFT and ReFT with two foundation models on MathQA$_\text{numeric}$ benchmark}
% \label{tab:mathqa_numeric}
% \end{table}

\paragraph{Reward Hacking for MathQA}
Our investigation of the negative results on MathQA$_\text{MCQ}$ indicates that ReFT suffers from the reward hacking~\cite{skalse2022defining} on the multi-choice question during training. 
Figure \ref{fig:reward_hacking} shows how the sampled solutions produce ``\textit{inaccurate rewards}'', which makes the RL training suffer. 
As we can see, the sampled CoT obtains an incorrect answer ``\textit{172}'' which is not half of the product of ``\textit{18}'' and ``\textit{22}''.
However, the final reasoning step still predicts the option ``\textit{C}'' as the final answer as the model would always predict one of the options from $\{\texttt{A, B, C, D, E}\}$ regardless of the correctness of intermediate CoT\footnote{We found that program-based CoTs are less likely to suffer as it is more rigorous than natural language.}. 
Thus, such a misleading CoT will receive a positive reward ``$1$'' and misguide the model to treat this as a correct CoT. 
The underlying reward hacking phenomenon severely tampers the model training~\cite{everitt2021reward}. 
This is also the reason that we chose the checkpoint with longer warm-up steps for MathQA N-CoT to reduce the reward hacking effect.
% https://arxiv.org/pdf/1908.04734.pdf
% model just learn to predict whatever ABCD option
% also the reason why we need to use later ckpt

\begin{table}[t!]
    \centering
    \adjustbox{max width=1.0\linewidth}{
    \begin{tabular}{lcccc}
    \toprule
         \multirow{2}{*}{\bf Method}& \multirow{2}{*}{\bf Size} &  \multicolumn{2}{c}{\bf GSM8K} & {\bf Extra SFT} \\
         & & \bf N-CoT &\bf  P-CoT & {\bf Data}\\
         \midrule
         % Galactica + SFT & ${41.0}$ & ${57.1}$   \\
         % Galactica + ReFT & ${46.8}$ & ${68.4}$   \\
         Galactica + SFT + Voting & 6.7B& $52.8$ & $62.9$  & \includegraphics[width=0.15in]{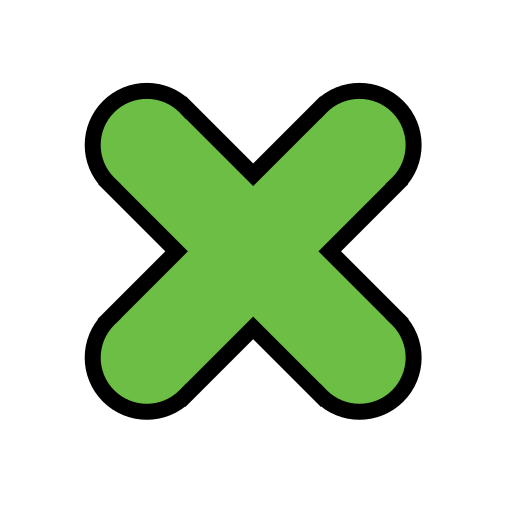}  \\
         Galactica + ReFT + Voting & 6.7B& $58.5$ & $71.8$ & \includegraphics[width=0.15in]{green_cross.png} \\
         \hdashline
         % \midrule
         Galactica + SFT + Reranking& 6.7B &  $57.5$ & $73.4$ & \includegraphics[width=0.15in]{green_cross.png} \\
         Galactica + ReFT + Reranking & 6.7B&  $\mathbf{59.2}$ & $\mathbf{76.4}$ & \includegraphics[width=0.15in]{green_cross.png} \\
         \midrule
         \midrule
         CodeLLAMA + SFT  + Voting & \textcolor{white}{0.}7B& $53.5$ & $68.0$ & \includegraphics[width=0.15in]{green_cross.png}\\
         CodeLLAMA + ReFT  + Voting & \textcolor{white}{0.}7B& $63.2$ & $78.0$ & \includegraphics[width=0.15in]{green_cross.png}\\
         % \midrule
         \hdashline
         CodeLLAMA + SFT  + Reranking& \textcolor{white}{0.}7B &  $62.9$ & $77.0$& \includegraphics[width=0.15in]{green_cross.png}\\
         % CodeLLAMA + ReFT  &${53.5}$ & ${72.7}$ \\
         CodeLLAMA + ReFT  + Reranking & \textcolor{white}{0.}7B&  $\mathbf{66.0}$ & $\mathbf{81.2}$ & \includegraphics[width=0.15in]{green_cross.png} \\
         \midrule
         \midrule
         \multicolumn{3}{l}{\textcolor{mygray}{Other Foundation Models}~$\dagger$} &\\
         \textcolor{black}{~WizardMath}~\cite{luo2023wizardmath} & \textcolor{white}{0}7B & $54.9$ & - & \includegraphics[width=0.15in]{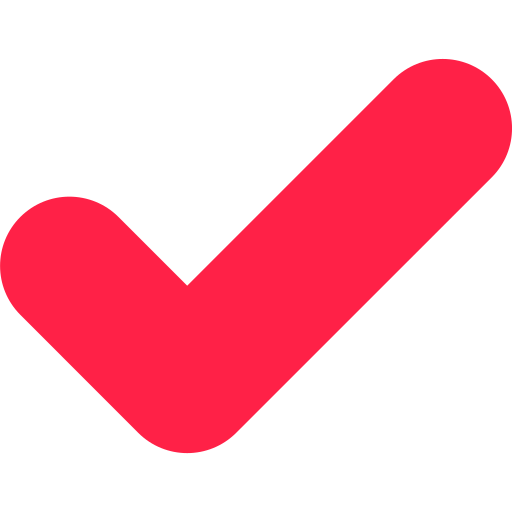} ($\textcolor{white}{0}96$k) \\
         \textcolor{black}{~WizardMath}~\cite{luo2023wizardmath} & 13B & $63.9$ & - & \includegraphics[width=0.15in]{check.png} ($\textcolor{white}{0}96$k)\\
         \textcolor{black}{~MathCoder}~\cite{wang2023mathcoder} & \textcolor{white}{0}7B & $67.8$ & - & \includegraphics[width=0.15in]{check.png} ($\textcolor{white}{0}80$k)\\ 
         \textcolor{black}{~MAmmoTH-Coder}~\cite{yue2023mammoth} & \textcolor{white}{0}7B & $22.2$ & $58.8$  & \includegraphics[width=0.15in]{check.png} ($260$k)\\
         \textcolor{black}{~MAmmoTH-Coder}~\cite{yue2023mammoth} & 70B & $72.4$ & $76.7$  & \includegraphics[width=0.15in]{check.png} ($260$k)\\
         \textcolor{black}{~DeepSeekMath}~\cite{shao2024deepseekmath} & \textcolor{white}{0}7B & $88.2$ & $86.7$ & \includegraphics[width=0.15in]{check.png} ($776$k)\\
         \midrule
         \midrule
         \textcolor{black}{~GPT-3.5-turbo~\cite{jie2023design}} & N.A. & $75.3$ & $78.0$& N.A.  \\
         \textcolor{black}{~GPT-4~\cite{openai2023gpt4, zhou2023solving}} & N.A. & $93.0$ & $97.0$  & N.A.\\
         \bottomrule
    \end{tabular}
    }
    % \vspace*{-3mm}
    \caption{Solving accuracy of majority voting and reward model reranking for SFT and ReFT on GSM8K. We also include existing approaches for comparison.}
    \label{tab:voting_reranking}
\end{table}

To further demonstrate the negative effect of MCQ questions, we experiment on the MathQA variant by \citet{jie2023leveraging}, MathQA$_\text{numeric}$ (Table \ref{tab:data_statistics}), which removes the options in the question, and directly predicts the numeric answer. 
% Table \ref{tab:mathqa_numeric} presents the comparison against SFT. 
% We can observe that ReFT consistently outperforms SFT using both Galactica and CodeLLAMA. 
Table \ref{tab:mathqa_numeric} presents the comparison against the baselines. 
We can observe that ReFT consistently outperforms the baselines using both Galactica and CodeLLAMA. 
% We use the same procedure and hyper-parameter in \S\ref{sec:setup} except that SFT checkpoint at 10th epoch is used to initialize ReFT. 
Ideally, we could reduce the reward hacking effect on MathQA$_\text{MCQ}$ if we can obtain a more fine-grained reward (e.g., process-based reward~\cite{lightman2023lets}) for the intermediate reasoning steps. 
However, the development of a reliable process-based reward model is expensive, and requires extensive manual annotations of reasoning steps. 
Recognizing these challenges, we consider controlling reward hacking and its analysis as an important problem to be addressed in future work.

\paragraph{Majority Voting and Reranking Benefit ReFT}
Following \citet{wang2022self,uesato2022solving,lightman2023lets}, we also perform majority voting and reward model reranking to show that ReFT can benefit from these common techniques. 
Specifically, we perform sampling from both SFT and ReFT policies. 
We sample $100$ CoT solutions for each question and employ the reward model described in \S\ref{sec:setup} to perform reranking. 
Results in Table \ref{tab:voting_reranking} demonstrate that ReFT consistently achieves the best performance on GSM8K by reward model reranking. 
ReFT + Voting significantly outperforms SFT + Voting by $8.6$ points on average across all settings.
ReFT with reranking outperforms SFT with reranking by more than $3$ points.
% on average. 

% 

Compared with existing open-source approaches~\cite{luo2023wizardmath,wang2023mathcoder,yue2023mammoth} (Table \ref{tab:voting_reranking} bottom\footnote{Numbers are taken from original papers. The N-CoT and P-CoT results for MAmmoTH-Coder are reported in their appendix.}), our best P-CoT variant achieves the best performance with accuracy $81.2$ on GSM8K. 
In addition, these approaches mainly include extra data generated from ChatGPT and perform distillation during fine-tuning. 
In contrast, we improve the policy itself by exploiting the potential of existing training data and pushing the limit of the policy performance. 
Our best result reported in Table \ref{tab:voting_reranking}, 
i.e., the CodeLLAMA + ReFT  + Reranking with P-CoT setting, 
even surpasses \texttt{GPT-3.5-turbo}.
However, we obtain the result with a model that is only in the size of 7B. 
\begin{table}[t!]
    \centering
     \adjustbox{max width=1.0\linewidth}{
    \begin{tabular}{lccc}
        \toprule
         \textbf{Method} &  \bf GSM8K & \bf SVAMP &\bf  MathQA$_\text{MCQ}$ \\
         \midrule
         Galactica-125M + SFT  & $23.7$ & $35.6$ & $58.4$ \\
         Galactica-125M + ReFT & $29.8$ & $39.4$ & $60.7$ \\
         \midrule
         Codeparrot-small + SFT  & $13.8$ & $25.7$ & $55.3$ \\
         Codeparrot-small + ReFT & $16.8$ & $27.4$ & $58.3$ \\
         \midrule
         Codegen-350M + SFT  & $20.4$ & $34.4$ & $56.4$ \\
         Codegen-350M + ReFT & $28.4$ & $39.3$ & $59.1$ \\
         \bottomrule
    \end{tabular}
    }
    % \vspace*{-3mm}
    \caption{Experiments on P-CoT with Galactica-125M, Codeparrot-small and Codegen-350M.}
    % \caption{Performance of Galactica-125M on P-CoT data.}
    \label{tab:small_model}
\end{table}

\begin{figure*}[t!]
    \centering
    \begin{subfigure}[b]{0.32\linewidth}
		\centering
		\adjustbox{max width=1\textwidth}{
			\begin{tikzpicture}
        	\tikzstyle{every node}=[font=\bfseries]
        	\begin{axis}[
        		xlabel style={font=\bfseries},
        		ylabel style={font=\bfseries},
        		width=10cm,
        		height=6cm,
        		grid=major,
        		grid style={dashed,gray!30},
        		enlarge x limits=false,
        		xmin=0, xmax=60, % Set the range for the x-axis
        		tick label style={font=\large, /pgf/number format/fixed},
        		axis line style = thick,
        		label style={font=\large},
        		legend style={font=\small, at={(1,0.3)}, anchor=north east, line width=0.4mm},
        		legend cell align={left},
        		cycle list name=exotic,
        		line width = 0.2mm,
        		no markers,
        		smooth
        		]
        		
        		% Load the data from an external file
        		\addplot table[x expr=\thisrowno{0}/460, y expr=\thisrowno{1}] {actual_reward.txt};
        		\addlegendentry{Actual Data}
        		\addplot+[draw=puffin] table[x expr=\thisrowno{0}, y expr=\thisrowno{1}] {reward_sma.txt};
        		\addlegendentry{10-point Smooth Moving Average}
        		% If you want to add a smooth moving average, you would need to calculate it
        		% and have it in a separate data file or add it manually as a new \addplot
        		
        	\end{axis}
        \end{tikzpicture}
		}
    \vspace*{-6mm}
    \caption{Mean Training reward}
	\end{subfigure}
     \begin{subfigure}[b]{0.32\linewidth}
     \centering
     \adjustbox{max width=1\textwidth}{
        \begin{tikzpicture}
        	\tikzstyle{every node}=[font=\bfseries]
        	\begin{axis}[
        		xlabel style={font=\bfseries},
        		ylabel style={font=\bfseries},
        		width=10cm,
        		height=6cm,
        		grid=major,
        		grid style={dashed,gray!30},
        		enlarge x limits=false,
        		xmin=0, xmax=300, % Set the range for the x-axis
        		tick label style={font=\large, /pgf/number format/fixed},
        		axis line style = thick,
        		label style={font=\large},
        		legend style={font=\small, at={(1,0.2)}, anchor=north east, line width=0.4mm},
        		legend cell align={left},
        		cycle list name=exotic,
        		line width = 1mm,
        		no markers,
        		smooth
        		]
        		
        		\addplot+[line width=2pt, draw=deeppurple] table[x expr=\thisrowno{0} / 460, y expr=\thisrowno{1}] {eval_accuracy_new.txt};

        	\end{axis}
        \end{tikzpicture}

         }
        \vspace*{-6mm}
    \caption{Evaluation accuracy}
     \end{subfigure}
     \begin{subfigure}[b]{0.32\linewidth}
     \centering
     \adjustbox{max width=1\textwidth}{
        \begin{tikzpicture}
        	\tikzstyle{every node}=[font=\bfseries]
        	\begin{axis}[
        		xlabel style={font=\bfseries},
        		ylabel style={font=\bfseries},
        		width=10cm,
        		height=6cm,
        		grid=major,
        		grid style={dashed,gray!30},
        		enlarge x limits=false,
        		xmin=0, xmax=60, % Set the range for the x-axis
        		tick label style={font=\large, /pgf/number format/fixed},
        		axis line style = thick,
        		label style={font=\large},
        		legend style={font=\small, at={(1,0.7)}, anchor=north east, line width=0.4mm},
        		legend cell align={left},
        		cycle list name=exotic,
        		line width = 0.2mm,
        		no markers,
        		smooth
        		]
        		
        		% Load the data from an external file
        		\addplot+[draw=darkgrass] table[x expr=\thisrowno{0}/460, y expr=\thisrowno{1}] {kl.txt};
        		% \addlegendentry{Mean Seq KL}
        		
        	\end{axis}
        \end{tikzpicture}
         }
         \vspace*{-6mm}
    \caption{Mean Sequence KL}
     \end{subfigure}
     \vspace*{-2mm}
    \caption{Training reward of ReFT, evaluation accuracy, KL against training epoch on GSM8K P-CoT. }
    \label{fig:training_reward}
\end{figure*}

\paragraph{Experiments with Small Model}
Intuitively, exploration could lead to imperfect demonstration with a small language model. 
We conduct an experiment on P-CoT data using Galactica-125M\footnote{\rurl{huggingface.co/facebook/galactica-125m}}, Codeparrot-small\footnote{\rurl{huggingface.co/codeparrot/codeparrot-small}} and Codegen-350M\footnote{\rurl{huggingface.co/Salesforce/codegen-350M-mono}}.
Table \ref{tab:small_model} shows the performance comparison between SFT and ReFT. 
Surprisingly, ReFT still outperforms SFT on three datasets.
% even with a small model. 
Such improvements demonstrate the robustness of ReFT during the exploration of reasonable programs.

% \begin{table}[t!]
%     \centering
%      \adjustbox{max width=0.9\linewidth}{
%     \begin{tabular}{lc}
%     \toprule
%          \textbf{Model Setting} &  \textbf{Accuracy}\\
%          \midrule
%         CodeLLAMA + ReFT & $72.7$ \\
%         ~~~~~~~  -- remove partial reward & $70.9$ \\
%         ~~~~~~~  -- KL coefficient $\beta = 0$ & \textit{collapse} \\
%         ~~~~~~~  -- non-shared value model & $72.6$ \\
%     \bottomrule
%     \end{tabular}
%     }
%     \caption{Ablation study on GSM8K P-CoT.}
%     \label{tab:ablation}
% \end{table}

\begin{table}[t!]
    \centering
     \adjustbox{max width=1.0\linewidth}{
    \begin{tabular}{lc}
    \toprule
         \textbf{Model Setting} &  \textbf{Accuracy}\\
         \midrule
        CodeLLAMA + ReFT & $75.28$ \\
        ~~~~~~~  -- remove partial reward & $74.40$ \\
        ~~~~~~~  -- KL coefficient $\beta = 0$ & \textit{collapse} \\
        ~~~~~~~  -- non-shared value model & $75.15$ \\
    \bottomrule
    \end{tabular}
    }
    % \vspace*{-3mm}
    \caption{Ablation study on GSM8K P-CoT.}
    \label{tab:ablation}
\end{table}

\paragraph{Ablation Study}
We perform the ablation study using CodeLLAMA on GSM8K P-CoT (Table \ref{tab:ablation}). 
Without the partial reward, ReFT obtains a lower accuracy $74.4$ but it is still much better than SFT. 
As mentioned in \S\ref{sec:reft}, such a partial reward can help reduce the effect of sparse reward~\cite{trott2019keeping} during training.
In addition, the policy distribution will easily collapse to produce unexpected results (i.e., $0$ accuracy) if we set the KL coefficient $\beta$ to $0$. 
It is certainly critical to impose constraints on the space that the policy explores~\cite{ouyang2022training}. 
The initial warm-up step essentially makes such constraints and allows the policy to further explore within the range that is governed by $\beta$. 
%Finally, we also experiment with a value model that has no parameter shared with the policy model~\cite{andrychowicz2021matters,cobbe2021phasic}. 
%The individual value model initializes the parameters the same as the policy model. 
%We found that such a setting allows the model to converge faster and eventually reach equivalent performance but sacrifices two times of original computation overhead as we have to perform the forward pass twice for each batch. 
We also experiment with a separate value model~\cite{andrychowicz2021matters,cobbe2021phasic}, 
where the torso parameters are initialized the same as the policy model.
We found that such a setting allows the policy to converge faster in early RL training, but eventually reaches an on par performance.
Compared to the original setting of a shared value model, 
it is, however, twice the computation overhead due to one extra forward-pass, 
as well as twice the memory cost due to the storage of the separate value net.
Finally, in Appendix~\ref{sec:case_study} we give a case study to show how the generated P-CoT evolve for SFT and ReFT.

\section{Analysis}
\label{sec:analysis}

\paragraph{Generalization}
% \label{sec:analysis_generalization}
Figure \ref{fig:training_reward} shows the mean reward, evaluation accuracy, and the KL divergence during training of ReFT\footnote{For illustration purpose, we only shows the mean reward and KL for $60$ epochs.} on GSM8K P-CoT using CodeLLAMA as foundation model. 
SFT converges and becomes overfiting when approaching 40$^{th}$ epoch. 
However, we can see the mean reward is around 80\% to 90\% for the ReFT policy at 40$^{th}$ epoch, and the value accuracy is also increasing. 
In addition, we can see that the KL divergence (Figure \ref{fig:training_reward} (c)) is very large in the beginning and then maintains a reasonable value between $0$ and $10$.
The stable KL divergence indicates our policy performs exploration within a space that contains appropriate programs. 
The underlying reinforcement learning mechanism greatly improves the generalization ability of ReFT~\cite{brown2020better}.

\paragraph{Qualitative Evaluation}
% To understand the response quality, we perform a human evaluation to assess the output from the SFT model, Warmup checkpoint, and ReFT model. The dataset used in our evaluation comprise of 50 questions in GSM8K testset that can be solved correctly by all three models. Their responses are assessed in three criteria: \textit{Logic}, \textit{Naming} and \textit{Compactness}. \textit{Logic} evaluates if the logic leading to the answer is correct,  \textit{Naming} evaluates if the variable conveys appropriate and reasonable semantics and \textit{Compactness} evaluates if the reasoning paths contain redundant information. Each is scored on a scale from 0 to 1 so the perfect score of 3 indicates good performance across these three dimensions. We employ four different annotators to score the reasoning path. To ensure the evaluation is impartial and faithful, we anonymize the origin of each reasoning path to prevent annotator bias.
% As seen in \ref{tab:qualitative_table}, though the overall scores are quite close (as the samples are all correctly predicted and program-based CoTs), ReFT performs slightly better than SFT, and outperforms the Warmup variant. Note that SFT is inherently trained to learn from the ground truth, and therefore it is likely to have a high score for the generated reasoning path. This comparative analysis underscores the robustness of ReFT in generating accurate and semantically coherent reasoning paths. 

We perform a human evaluation to qualitatively assess the output from the SFT model, Warmup checkpoint, and ReFT model. 
The evaluation uses 50 questions and samples the solutions in GSM8K test set that can be solved correctly by all three models. 
We ask four different annotators to score the reasoning path according to the following criteria, each scored on a scale from $0$ to $1$.
\squishlist
% The outputs are assessed in 3 dimensions with each scored on a scale of 0 to 1. 
\item \textit{Logic}: evaluates if the logic leading to the answer is correct.
\item \textit{Naming}: evaluates if the variable conveys appropriate and reasonable semantics 
\item \textit{Compactness}: evaluates if the reasoning paths contain redundant information. 
\squishend
A perfect score of 3 indicates good performance across these three dimensions.
To ensure the evaluation is impartial and faithful, we strictly follow the setting:
(1) The origin of each reasoning path (from SFT, Warmup, or ReFT) is anonymized to prevent annotator bias.
(2) Four different annotators are responsible for different portions of the samples.

\begin{table}[t!]
    \centering
    \adjustbox{max width=1.0\linewidth}{
        \begin{tabular}{lcccc}
        \toprule
        {\textbf{Method}} &  \textbf{Logic} &  \textbf{Naming} &  \textbf{Compactness} &  \textbf{Overall Score} \\ \midrule
        {\textbf{SFT}}    &  0.986          &  0.988           &  0.994                &  2.967                  \\ 
        {\textbf{Warmup}} &  0.949          &  0.982           &  0.990                &  2.920                  \\ 
        {\textbf{ReFT}}   &  0.992          &  0.990           &  0.996                &  \textbf{2.982}         \\ \bottomrule
        \end{tabular}
    }
    \caption{Qualitative scores of models from three methods trained on GSM8k P-CoT dataset.}
    \label{tab:qualitative_table}
\end{table}

As seen in table \ref{tab:qualitative_table}, though the overall scores are quite close, ReFT performs slightly better than SFT, and outperforms the Warmup variant. Note that SFT is inherently trained to learn from the ground truth, thus, it is likely to have a high score.
This comparative analysis underscores the robustness of ReFT in generating accurate and semantically coherent reasoning paths.

\paragraph{When ReFT surpasses SFT?}
% \subsection{When ReFT surpasses SFT?}
% \label{sec:reft_epoch}
To further investigate the relationship between ReFT and SFT, we perform ReFT training with different number of warm-up steps from SFT. 
Figure \ref{fig:reft_epoch} shows the value accuracy of different ReFT variants against SFT\footnote{We only show 60 epochs for illustration purposes. The performance for the later epoch is shown in Figure \ref{fig:training_reward} (b).}. 
Specifically, if the warmup step is $3$, that means the policy initialize from the $3^{rd}$-epoch SFT checkpoint. 
We can see that the performance of all ReFT policies decreases right after the warm-up in the beginning, until the training epoch reaches around $8$.
% all ReFT policies have worse performance in the beginning where the epoch is less than $8$.
Because the linear layer in the shared value model is randomly initialized, and it could take a few epochs to adjust the distribution. 
Starting from the $30^{th}$ epoch, SFT converges and all ReFT variants are still improving. 
We can also see that all variants outperform SFT by a significant margin and there is no obvious advantage of any specific ReFT variant.

%%%%% conclusion backup begins
% \section{Conclusion}
% Motivated by the existence of multiple reasonable CoTs, we proposed reinforced fine-tuning (ReFT), a new fine-tuning paradigm using reinforcement learning to solve math problems. 
% Compared with SFT, ReFT directly optimizes a non-differentiable objective where the policy will explore a certain number of CoT alternatives toward the correct answer rather than optimizing a single CoT. 
% Through extensive experiments on three datasets with two foundation models, we demonstrate that ReFT has superior performance and much better generalization ability than SFT. 
% Furthermore, we show that the underlying model trained by ReFT also benefits from common techniques such as majority voting~\cite{wang2022self} and reward model reranking~\cite{cobbe2021training,uesato2022solving}. 
% ReFT also outperforms several publicly open-source and strong models with the same scale.
% % a new fine-tuning paradigm, reinforced fine-tuning (ReFT) using reinforcement learning for solving math problems. 
% % propose new fine-tuning algorithm
% % advantages
% % better performance, benefits from common techniques
% % generalization, 
%%%%% conclusion backup ends

\begin{figure}[t!]
    \centering
    \adjustbox{max width=0.95\linewidth}{
    \begin{tikzpicture}
	\tikzstyle{every node}=[font=\bfseries]
	\begin{axis}[
		xlabel={Training Epoch},
		xlabel style={font=\bfseries},
		ylabel style={font=\bfseries},
		grid=major,
		grid style={dashed,gray!30},
		enlarge x limits=false,
		xmin=0, xmax=60, % Set the range for the x-axis
		tick label style={font=\large, /pgf/number format/fixed},
		axis line style = thick,
		label style={font=\large},
		legend style={font=\small, at={(1,0.5)}, anchor=north east, line width=0.4mm},
		legend cell align={left},
		cycle list name=exotic,
		line width = 1mm,
		no markers,
		smooth
		]
		
		\addplot+[line width=2pt,restrict x to domain=0:60] table[x expr=\thisrowno{0}, y expr=\thisrowno{3}] {reft_ep.txt};
		\addplot+[line width=2pt,restrict x to domain=0:60] table[x expr=\thisrowno{0}, y expr=\thisrowno{4}] {reft_ep.txt};
		\addplot+[line width=2pt,restrict x to domain=0:60] table[x expr=\thisrowno{0}, y expr=\thisrowno{5}] {reft_ep.txt};
		\addplot+[line width=2pt,restrict x to domain=0:60] table[x expr=\thisrowno{0}, y expr=\thisrowno{6}] {reft_ep.txt};
		\addplot+[line width=2pt,restrict x to domain=0:40] table[x expr=\thisrowno{0}, y expr=\thisrowno{1}] {reft_ep.txt};
        % \addplot+[line width=2pt,restrict x to domain=0:50] table[x expr=\thisrowno{0}, y expr=\thisrowno{2}] {reft_ep.txt};
        
		\addlegendentry{ReFT$_\text{warm-up\_ep.=1}$}
		\addlegendentry{ReFT$_\text{warm-up\_ep.=2}$}
		\addlegendentry{ReFT$_\text{warm-up\_ep.=3}$}
		\addlegendentry{ReFT$_\text{warm-up\_ep.=4}$}
		\addlegendentry{SFT}
        % \addlegendentry{ReFT$_\text{warmup\_step=0}$}
		% If you want to add a smooth moving average, you would need to calculate it
		% and have it in a separate data file or add it manually as a new \addplot
		
	\end{axis}
\end{tikzpicture}
    }
    \vspace*{-3mm}
    \caption{Accuracy comparison between SFT and ReFT with different number of warm-up epoch.}
    \label{fig:reft_epoch}
\end{figure}

%%%%% by hang
\section{Conclusion}
We have introduced reinforced fine-tuning (ReFT) as a new method for fine-tuning models to solve math problems. In contrast to SFT, ReFT optimizes a non-differentiable objective by exploring multiple CoT annotations in the search for the correct answer, rather than relying on a single annotation.
% single CoT annotation.

Through extensive experimentation on three datasets using two foundation models, we have demonstrated that ReFT outperforms SFT in terms of performance and generalization ability. Moreover, we have showcased the compatibility of models trained with ReFT with techniques such as majority voting~\cite{wang2022self} and reward model reranking~\cite{cobbe2021training,uesato2022solving}.

Furthermore, ReFT has exhibited superior performance compared to several publicly available open-source models of comparable sizes in math problem-solving. This demonstrates the effectiveness and practical value of the ReFT approach.

\section{Future Work}
% We made the first attempt to apply reinforcement learning, specifically PPO algorithm~\cite{schulman2017proximal}, for fine-tuning. 
% Future work includes applying offline reinforcement learning~\cite{levine2020offline,gulcehre2023reinforced} to further improve the performance and close the gap with reranking. 
% In addition, \citet{lightman2023lets} indicates that a well-trained process-based reward model (PRM) can greatly improve performance. 
% We would like to further explore the possibility with the help of PRM. 
% Finally, we would like to investigate the possibility for a \textit{warm-up free} approach as the warm-up step is essential for ReFT as we can see in \S\ref{sec:reft_epoch}. 
We have made the first attempt of applying reinforcement learning, specifically the PPO algorithm~\cite{schulman2017proximal}, to fine-tune of LLMs for math problem-solving. Our future work includes utilization of offline reinforcement learning techniques~\cite{levine2020offline,gulcehre2023reinforced}, development of a \textit{warm-up free} method to enhance training efficiency and performance, thereby reducing the gap with the reranking method. Additionally, \citet{lightman2023lets} suggests that a well-trained process-based reward model (PRM) can significantly enhance performance. Hence, it would be worthwhile to explore the implementation of process-based rewards in reinforcement learning training. 
Lastly, as ReFT is a versatile approach, we intend to apply it to more general reasoning tasks where the inference can be formalized with CoT.
% Finally, ReFT 
% Finally, we would like to investigate the possibility for a \textit{warm-up free} approach as the warm-up step is essential for ReFT as we can see in \S\ref{sec:reft_epoch}. 
% 
% no warmup required
% better reward model. 

\section*{Limitations}
\paragraph{Training Efficiency}
As depicted in Figure \ref{fig:training_reward} (b), it is evident that ReFT necessitates a greater number of epochs to reach convergence compared to SFT. This is primarily due to the fact that ReFT optimizes a non-differentiable objective and requires exploration of the generation space to attain correct answers. While a larger learning rate may expedite convergence, it also makes the policy more susceptible to instability and potential collapse. Alternatively, using a larger batch size is a viable option; however, it comes at the expense of increased computational costs.

\paragraph{Reward Hacking}
Our reward function relies solely on the final answer to determine the reward. However, as demonstrated in the experiments conducted on the MathQA$_\text{MCQ}$ N-CoT dataset, the policy can be easily manipulated if the possible space of final answers is limited, such as ${\texttt{A,B,C,D}}$. To mitigate the issue of reward hacking, it may be necessary to employ a more detailed or process-based reward function that takes into account a broader range of factors.

% \paragraph{Warm-up}
% reward hacking 
% 

\bibliography{custom}

\appendix

\section{Examples of N-CoT and P-CoT Representations}
\label{sec:appendix_representation}
We present examples of natural language CoT and program-based CoT from GSM8K dataset in Figure \ref{fig:cot_example}. 
We follow \citet{jie2023design} to perform few-shot prompting and obtain the CoT representations. 
The natural language CoT is generally the same as the one presented in \citet{wei2022chain}. 
The format program-based CoT is similar to the one in PAL~\cite{gao2023pal}, where we use a function to solve the problems. 
The function starts with a Python docstring that repeats the question and then a list of statements as reasoning steps.

\begin{figure}[t!]
    \centering
    \adjustbox{max width=1.0\linewidth}{
		\begin{tikzpicture}[node distance=2.0mm and 2.0mm, >=Stealth, 
			wordnode/.style={draw=none, minimum height=5mm, inner sep=0pt},
			chainLine/.style={line width=0.8pt,-, color=mygray},
			entbox/.style={draw=black, rounded corners, fill=red!20, dashed},
			mathop/.style={draw=none, circle, minimum height=2mm, inner sep=1pt, line width=0.8pt, fill=pinkgla},
			stepsty/.style={draw=pinkgla, circle, minimum height=2mm, inner sep=1pt, line width=0.8pt},
			quant/.style={draw=none, minimum height=2mm, inner sep=1pt, line width=1pt},
			expr/.style={draw=electronblue, rectangle, minimum width=20mm, minimum height=4.5mm, inner sep=2pt, line width=0.8pt, rounded corners},
			box/.style={draw=electronblue, rectangle, rounded corners, line width=0.9pt, dashed, minimum width=26mm, minimum height=20mm},
			invis/.style={draw=none},
            highlight/.style={fill=yellow!50, rounded corners}
			]

            \node[wordnode, align=left, text width=12cm, font=\fontfamily{phv}\selectfont] (question) [] {\textbf{Question}: Weng earns \$12 an hour for babysitting. Yesterday, she just did 50 minutes of babysitting. How much did she earn?};

            \node[wordnode, align=left, text width=12cm, font=\fontfamily{phv}\selectfont, below=of question, draw=electronblue, inner sep=3pt, line width=1.5pt] (cot) [] {\textbf{N-CoT}: We need to calculate her hourly rate and then multiply  it by the amount of time she worked. \\~\\
            First, we need to convert 50 minutes to hours. \\~\\
            There are 60 minutes in an hour, so 50 minutes is equal to 50/60 = 5/6 hours. \\~\\
            Next, we can calculate Weng's earnings by multiplying her hourly rate
by the amount of time she worked: \\~\\
            ~~~~\$12/hour x 5/6 hour = \$10.\\~\\
            Therefore, Weng earned \$10 for 50 minutes of babysitting. \\~\\
The answer is 10.};

            \node[wordnode, align=left, text width=12cm, font=\fontfamily{phv}\selectfont, below=of cot, draw=electronblue, inner sep=3pt, line width=1.5pt] (pot) [] {\textbf{P-CoT}: \\
            def solution():\\
            ~~~~"""Weng earns \$12 an hour for babysitting. Yesterday, she just did 50 minutes of babysitting. How much did she earn?"""\\
            ~~~~hourly\_rate = 12 \\
            ~~~~minutes\_worked = 50\\
            ~~~~hours\_worked = minutes\_worked / 60\\
            ~~~~earnings = hourly\_rate * hours\_worked \\
            ~~~~result = earnings \\
            ~~~~return result
            };

		\end{tikzpicture}
	}
    \caption{N-CoT and P-CoT examples on GSM8K}
    \label{fig:cot_example}
\end{figure}

\section{Detailed Hyperparameter Setting}
\label{sec:appendix_hyperparameter}

\paragraph{Supervised Fine-Tuning}
% We train the model for 40 epochs, the maximum length is set to $1024$.
% The batch size is 48. 
% Input with longer length is to be truncated to the first $1024$ tokens.
We train the model for 40 epochs with the batch size of 48 and the maximum length of $1024$..  For small models, we increase the learning rate to $2e$-$5$, and the number of epoch for training MathQA$_\text{MCQ}$ to 100 epochs.

\paragraph{ReFT Warm-up}
% For Galactica, we perform warm-up for 2 epochs on GSM8K, SVAMP for both N-CoT and P-CoT.
% In terms of MathQA$_\text{MCQ}$, we perform warm-up for 5 epochs on MathQA$_\text{MCQ}$ N-CoT and 2 epochs on MathQA$_\text{MCQ}$ P-CoT.
% For CodeLLAMA, we perform warm-up for 1 epoch on SVAMP, 2 epochs on GSM8K, 5 epochs on MathQA$_\text{MCQ}$ N-CoT and 2 epochs on MathQA$_\text{MCQ}$ P-CoT.
We perform warm-up for 2 epochs on GSM8K, SVAMP for both N-CoT and P-CoT.
For MathQA$_\text{MCQ}$, we perform warm-up for 5 epochs on MathQA$_\text{MCQ}$ N-CoT and 2 epochs on MathQA$_\text{MCQ}$ P-CoT.
Specifically for MathQA$_\text{numeric}$, we perform warm-up for 10 epochs because this dataset is much harder and the number of reasoning chains is longer than other datasets. 
% For small models, we the warm-up period is 10 epochs for GSM8K and SVAMP and is 40 epochs for MathQA$_\text{MCQ}$
For Galactica-125m and Codegen-350M, the warm-up period is 10 epochs for GSM8K and SVAMP and is 40 epochs for MathQA$_\text{MCQ}$. For Code-parrot, we increases the warm-up period to 40 epochs on all datasets to obtain reasonable warm-up performance.

\paragraph{ReFT RL}
The maximum length for question is set to $300$, and the maximum length during sampling is set to $700$. 
The batch size is 32, which is smaller than SFT due to extra memory consumption of the value model. 
The number of updates per RL step (i.e., ppo epoch) is set to 2~\cite{ziegler2019fine}.
We do not employ any weight decay and dropout following \citet{ziegler2019fine}.
% For small models, we train for 700 epochs with the learning rate of $3e$-$6$ and the global batch size of 256.
For small models, we train for 700 epochs with the learning rate of $3e$-$6$, the global batch size of 256 and the $\alpha$ of 5, 1 and 0.1 for Galactica-125m, Codeparrot-small and Codegen-350M model respectively.

% We train the model for 40 epochs and select one checkpoint with the training set top-1 accuracy between 50\%-80\% as output model of this phase. 
% With Galactica foundation model, 2nd epoch checkpoint is used for GSM8k, SVAMP and MathQA P-CoT and 5th epoch checkpoint is used for MathQA-CoT. 
% With CodeLlama foundation model, 1st epoch checkpoint is used for SVAMP, 2nd epoch checkpoint is used for GSM8k and MathQA.

\section{Case Study}
\label{sec:case_study}
% We show how SFT and ReFT evolve by investigating the generated P-CoT for a specific given question.
% As in Figure~\ref{fig:sft_rl_evolve}, the question is selected from GSM8K testing set. 
% The three rows correspond to training epoch 1, 3, 5, respectively. 
% As seen in Figure~\ref{fig:sft_rl_evolve}, at epoch 1, ReFT is in warmup stage so that its generated P-CoT looks exactly the same with SFT. 
% However, later at epoch 3 and 5, the P-CoTs differ. 
% We can see that at epoch 5, ReFT begins to generate correct P-CoT, while SFT still struggles.

We show how SFT and ReFT evolve by investigating the generated P-CoT for a specific question.
Figure~\ref{fig:sft_rl_evolve} reports the responses of SFT and ReFT at checkpoint epoch 1, 3 and 5.
At epoch 1, ReFT is in warmup stage so that its generated P-CoT looks similar to that of SFT. 
However, later at epoch 3 and 5, the P-CoTs differ. ReFT responses becomes shorter and correct while SFT reponses remains incorrect.

% \begin{figure*}
%     \centering
%     \adjustbox{max width=1.0\linewidth}{
%         \includegraphics{sft_reft_evolve.jpeg}
%     }
%     \vspace*{-5mm}
%     \caption{Compare how SFT and RL evolve. See the text for more explanations.}
%     \label{fig:sft_rl_evolve}
% \end{figure*}

\section{Attempts with DPO and IPO}
% In addition to the PPO algorithm in this work, we initialliy experimented with  DPO~\cite{rafailov2023direct} and IPO~\cite{azar2023general} in our early attempts on the GSM8K dataset. 
% We obtained the preference data by sampling from the warmup checkpoint and adapted the implementation of these algorithms from the Transformer Reinforcement Learning (TRL) framework~\cite{vonwerra2022trl}.
% % adapted the implementation of these algorithms from the Transformer Reinforcement Learning (TRL) framework~\cite{vonwerra2022trl}.
% Preliminary results indicate that their performance is on par with the "Offline Self-Learning" baseline on GSM8K.
% % The preliminary experiments suggest the performance is comparable with the ``Offline Self-Learning'' baseline on GSM8K. 
% However, PPO is more suitable for the task of mathematical problem solving, as constructing positive and negative training pairs for DPO and IPO is not straightforward.

%%%%%%% 

In addition to the PPO algorithm in this work, we initially experimented with  DPO~\cite{rafailov2023direct} and IPO~\cite{azar2023general} in our early attempts on the GSM8K dataset. 
% We obtained the preference data by samples responses for questions in the training set using the warmup checkpoint and randomly pairs the correct and incorrect responses.
% Our implementation is adapted from the Transformer Reinforcement Learning (TRL) framework~\cite{vonwerra2022trl}.
We obtained the preference data by sampling from the warmup checkpoint and adapted the implementation of these algorithms from the Transformer Reinforcement Learning (TRL) framework~\cite{vonwerra2022trl}.
Preliminary results indicate that their performance is on par with the "Offline Self-Learning" baseline on GSM8K. 
This could be explained by the following reasons. 
% As DPO/IPO are an offline methods, their performances are limited by the quality of data sampled from an sub-optimal policy as they cannot explore more CoT paths by themselves.
Being offline methods, DPO and IPO cannot explore more CoT paths by themselves, thus, their performances are limited by the quality of data sampled from an sub-optimal policy.
% obtained from a suboptimal checkpoint. 
% Neither DPO nor IPO can explore more CoT solution paths by itself.
% In addition, since their policies are set to maximize the reward given by their implicitly trained reward models, having quality data for reward modeling is critical. However, our preference data is obtained by sampling from the training set, it could be imbalanced. For example, there may be neither incorrect responses for easy questions nor correct responses for hard questions. 
In addition, their policies are set to maximize the reward given by their implicitly trained reward models. Our preference data, which is sampled from the train set, might not be ideal for reward modeling as there may be neither incorrect responses for easy questions nor correct responses for hard questions. 
On the other hand, PPO not only is an online method that explores CoT solution paths automatically, but it also does not require reward model training as the reward function can be directly used. Thus, PPO is chosen in our later experiments.

\begin{figure*}[]
    \begin{tcbitemize}[raster columns=1, raster equal height=rows, size=small]
        \tcbitem[title=Example from GSM8K,colback=white]
            \begin{lstlisting}[style=plain]
Question: 
If Sam and Harry have 100 feet of fence between them, and they agree to split it with Harry getting 60 feet more than Sam, how much is left over for Sam?
        \end{lstlisting}
% Answer: 20
    \end{tcbitemize}
    \begin{tcbitemize}[raster columns=2, raster equal height=rows, size=small]
        \tcbitem[title=SFT,colback=white, colframe=red!50!black]
        \begin{lstlisting}[style=python, rulecolor=\color{red}, frame=single]
# Epoch 1
def solution():
    fence_initial = 100
    fence_difference = 60
    fence_harry = fence_initial + fence_difference
    fence_sam = fence_harry - fence_difference
    fence_left = fence_initial - fence_sam
    result = fence_left
    return result
        \end{lstlisting}
        \begin{lstlisting}[style=python, rulecolor=\color{red}, frame=single]
# Epoch 3
def solution():
    fence_initial = 100
    harry_feet = 60
    sam_feet = harry_feet - 1
    sam_feet = sam_feet * 2
    sam_feet = sam_feet / 2
    sam_feet = sam_feet
    result = sam_feet
    return result
        \end{lstlisting}
    
        \begin{lstlisting}[style=python, rulecolor=\color{red}, frame=single]
# Epoch 5
def solution():
    fence_length = 100
    harry_extra_feet = 60
    harry_feet = fence_length / 2 + harry_extra_feet
    sam_feet = fence_length / 2
    left_over_feet = fence_length - harry_feet - sam_feet
    result = left_over_feet
    return result
        \end{lstlisting}
        \tcbitem[title=ReFT,colback=white,colframe=green!50!black]
        \begin{lstlisting}[style=python, rulecolor=\color{red}, frame=single]
# Epoch 1
def solution():
    fence_total = 100
    fence_harry = fence_total - 60
    fence_sam = fence_harry - 60
    fence_left = fence_total - fence_harry - fence_sam
    result = fence_left
    return result
        \end{lstlisting}
        \begin{lstlisting}[style=python, rulecolor=\color{red}, frame=single]
# Epoch 3
def solution():
    fence_total = 100
    sam_fence = (fence_total / 2) - 60
    result = sam_fence
    return result
        \end{lstlisting}
    
        \begin{lstlisting}[style=python, rulecolor=\color{green}, frame=single]
# Epoch 5
def solution():
    fence_total = 100
    sam_fence = (fence_total - 60) / 2
    result = sam_fence
    return result
        \end{lstlisting}
    \end{tcbitemize}
% \caption{Compare how SFT and RL evolve. See the text for more explanations.}
\caption{P-CoT responses of SFT and ReFT checkpoints at epoch 1,3 and 5 to the same question in GSM8K dataset. Reponses in green frame are correct while responses in red frame are incorrect.}
\label{fig:sft_rl_evolve}
\end{figure*}

\end{document}